\documentclass[10pt,twocolumn,letterpaper]{article}

\usepackage[pagenumbers]{cvpr} 

\usepackage{graphicx}
\usepackage{amsmath}
\usepackage{amssymb}
\usepackage{booktabs}
\usepackage{pifont}
\newcommand{\cmark}{\ding{51}}%
\newcommand{\xmark}{\ding{55}}%
\usepackage{stmaryrd}

\usepackage{algorithm}
\usepackage{algpseudocode}
\usepackage{amsmath}
\DeclareMathOperator*{\argmax}{arg\,max}

\usepackage{mathtools}
\usepackage[textsize=tiny]{todonotes}
\usepackage[accsupp]{axessibility}

\usepackage[pagebackref,breaklinks,colorlinks]{hyperref}

\usepackage[capitalize]{cleveref}
\crefname{section}{Sec.}{Secs.}
\Crefname{section}{Section}{Sections}
\Crefname{table}{Table}{Tables}
\crefname{table}{Tab.}{Tabs.}

\def\confName{CVPR}
\def\confYear{2023}

\begin{document}
\title{Diffuse, Attend, and Segment: Unsupervised Zero-Shot Segmentation using Stable Diffusion}

\author{Junjiao Tian \\{jtian73@gatech.edu}\\
\and
Lavisha Aggarwal\\{lavishaggarwal@google.com}\\
\and
Andrea Colaco\\{andreacolaco@google.com}\\
\and 
Zsolt Kira\\{zk15@gatech.edu}\\
\and
Mar Gonzalez-Franco\\{margon@google.com}\\
}

\maketitle

\begin{abstract}
Producing quality segmentation masks for images is a fundamental problem in computer vision. Recent research has explored large-scale supervised training to enable zero-shot transfer segmentation on virtually any image style and unsupervised training to enable segmentation without dense annotations. However, constructing a model capable of segmenting anything in a zero-shot manner without any annotations is still challenging. In this paper, we propose to utilize the self-attention layers in stable diffusion models to achieve this goal because the pre-trained stable diffusion model has learned inherent concepts of objects within its attention layers. Specifically, we introduce a simple yet effective iterative merging process based on measuring KL divergence among attention maps to merge them into valid segmentation masks. The proposed method does not require any training or language dependency to extract quality segmentation for any images. On COCO-Stuff-27, our method surpasses the prior unsupervised zero-shot transfer SOTA method by an absolute $26\%$ in pixel accuracy and $17\%$ in mean IoU. The project page is at~\url{https://sites.google.com/view/diffseg/home}.
\end{abstract}

\section{Introduction}
\label{sec:intro}
\begin{figure}[ht]
    \centering
    \includegraphics[width=0.35\textwidth]{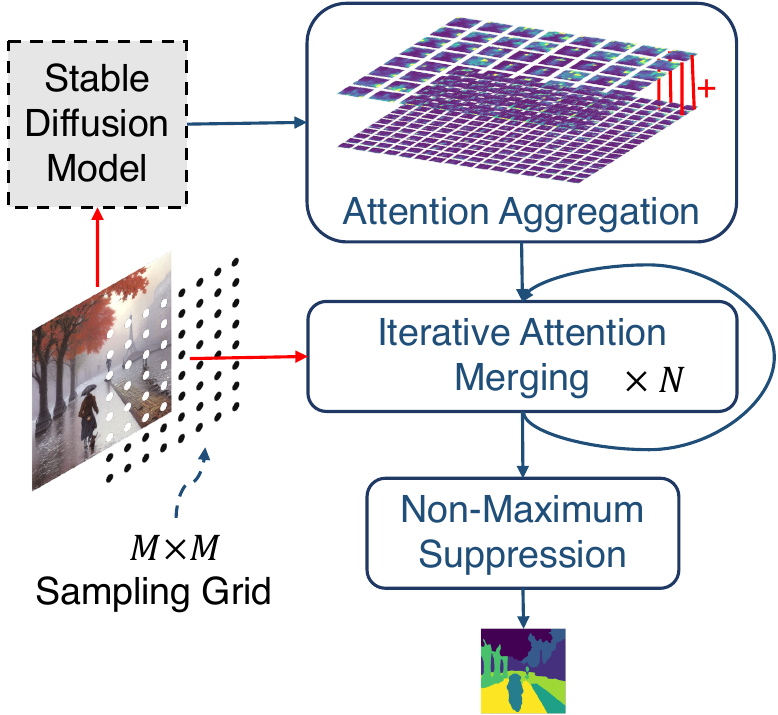}
    \caption{Overview of DiffSeg. DiffSeg is an unsupervised and zero-shot segmentation algorithm using a pre-trained stable diffusion model. Starting from $M\times M$ anchor points, DiffSeg iteratively merges self-attention maps from the diffusion model for $N$ iterations to segment any image without any prior knowledge and external information. }
    \label{fig:overview}
\end{figure}
Semantic segmentation divides an image into regions of entities sharing the same semantics. It is an important foundational application for many downstream tasks such as image editing~\cite{ling2021editgan}, medical imaging~\cite{liu2021review}, and autonomous driving~\cite{feng2020deep} etc. While supervised semantic segmentation, where a target dataset is available and the categories are known, has been widely studied~\cite{xie2021segformer,wang2021max,chen2019rethinking}, zero-shot transfer segmentation\footnote{\color{black}Zero-shot transfer segmentation means that the underlying model is not trained for segmentation, similar to CLIP which is not trained for classification but can be used for zero-shot classification~\cite{radford2021learning}.} for any images with unknown categories is much more challenging. A recent popular work SAM~\cite{kirillov2023segment}, trains a neural network on 1.1B segmentation annotations and achieves impressive zero-shot transfer to any images. This is an important step towards making segmentation a more flexible foundation task for many other tasks, not just limited to a given dataset with limited categories. However, the cost of collecting per-pixel labels is high. Therefore, it is of high research and production interest to investigate \textit{unsupervised} and \textit{zero-shot transfer} segmentation methods under the least restrictive settings, i.e., no form of annotations and no prior knowledge of the target.

Few works have taken up the challenge due to the combined difficulty of the unsupervised and zero-shot requirements. For example, most works in unsupervised segmentation require access to the target data for unsupervised adaption~\cite{feng2023network,zadaianchuk2022unsupervised,li2023acseg,shin2023namedmask,hamilton2022unsupervised,cho2021picie,ji2019invariant}. Therefore, these methods cannot segment images that are not seen during the adaptation. Recently, ReCo~\cite{shin2022reco} proposed a retrieve-and-co-segment strategy. It removes the requirement for supervision because it uses an unsupervised vision backbone, e.g., DINO~\cite{caron2021emerging}, and is zero-shot since it does not need training on the target distribution of images. Still, to satisfy these requirements, ReCo needs to 1) identify the concept beforehand and 2) maintain a large pool of unlabelled images for retrieval. This is still far from the capability of SAM~\cite{kirillov2023segment}, which does not require knowledge of the target image and any auxiliary information. 

To move towards the goal, we propose to utilize the power of a stable diffusion (SD) model~\cite{rombach2022high} to construct a general segmentation model. Recently, stable diffusion models have been used to generate prompt-conditioned high-resolution images~\cite{rombach2022high}. It is reasonable to hypothesize that there exists information on object groupings in a diffusion model. For example, DiffuMask~\cite{wu2023diffumask} discovers that the \textit{cross-attention} layer contains explicit object-level pixel groupings when cross-referenced between the produced attention maps and the input prompt. However, DiffuMask can only produce segmentation masks for a generated image with a corresponding text prompt and is limited to dominant foreground objects explicitly referred to by the prompt. Motivated by this, we investigate the unconditioned \textit{self-attention} layer in a diffusion model. We observed that the attention tensors from the self-attention layers contain specific spatial relationships: \textbf{Intra-Attention Similarity} and \textbf{Inter-Attention Similarity} (Sec.~\ref{sec:diffusion_model}). Relying on these two properties, it is possible to uncover segmentation masks for any objects in an image.

To this end, we propose DiffSeg (Sec.~\ref{sec:diffsam}), a simple yet effective post-processing method, for producing segmentation masks by using the attention tensors generated by the self-attention layers in a diffusion model solely. The algorithm consists of three main components: attention aggregation, iterative attention merging, and non-maximum suppression, as shown in Fig.~\ref{fig:overview}. Specifically, DiffSeg aggregates the 4D attention tensors in a spatially consistent manner to preserve visual information across multiple resolutions and uses an iterative merging process by first sampling a grid of anchor points. The sampled anchors provide starting points for merging attention masks, where anchors in the same object are absorbed eventually. The merging process is controlled by measuring the similarity between two attention maps using KL divergence. Unlike, popular clustering-based unsupervised segmentation methods~\cite{cho2021picie,hamilton2022unsupervised,li2023acseg}, DiffSeg does not require specifying the number of clusters beforehand and is also deterministic. Given an image, without any prior knowledge, DiffSeg can produce a quality segmentation without resorting to any additional resources (e.g. as SAM requires). We benchmark DiffSeg on the popular unsupervised segmentation dataset COCO-Stuff-27 and a specialized self-driving dataset Cityscapes. DiffSeg outperforms prior works on both datasets despite requiring less auxiliary information during inference. In summary, 
\begin{itemize}
    \item We propose an unsupervised and zero-shot segmentation method, DiffSeg, using a pre-trained stable diffusion model.  DiffSeg can segment images in the wild without any prior knowledge or additional resources.
    \item DiffSeg sets state-of-the-art performance on two segmentation benchmarks. On COCO-Stuff-27, our method surpasses a prior unsupervised zero-shot SOTA method by an absolute $26\%$ in pixel accuracy and $17\%$ in mean IoU.
\end{itemize}
\section{Related Works}
\label{sec:related_works}
\textbf{Diffusion Models.}
Our work builds on pre-trained stable diffusion models~\cite{ho2020denoising,rombach2022high,saharia2022photorealistic}. Many existing works have studied the discriminative visual features in them and used them for zero-shot classification~\cite{li2023your},  supervised segmentation~\cite{amit2021segdiff}, label-efficient segmentation~\cite{baranchuk2021label}, semantic-correspondence matching~\cite{hedlin2023unsupervised,zhang2023tale} and open-vocabulary segmentation~\cite{xu2023open}. They rely on the high-dimensional \textit{visual features} learned in stable diffusion models to perform those tasks requiring additional training to take full advantage of those features. Instead, we show that object grouping is also an emergent property of the \textit{self-attention} layer in the Transformer module manifested in 4-dimensional spatial attention tensors with particular spatial patterns.

\textbf{Unsupervised Segmentation.} Our work is closely related to unsupervised segmentation~\cite{feng2023network,zadaianchuk2022unsupervised,li2023acseg,shin2023namedmask,hamilton2022unsupervised,cho2021picie,ji2019invariant}. This line of works aims to generate dense segmentation masks without using any annotations. However, they generally require unsupervised training on the target dataset to achieve good performance. We characterize the most recent works into two main categories: clustering based on invariance~\cite{cho2021picie,ji2019invariant} and, more recently, clustering using pre-trained models~\cite{hamilton2022unsupervised,li2023acseg,zadaianchuk2022unsupervised,nguyen2019deepusps,wu2023diffumask}. 
Our work falls in the second category. To produce segmentation masks, these new works utilize the discriminative features learned in those pre-trained models, i.e., features from the same semantic class are generally more similar. DiffuMasks~\cite{wu2023diffumask} uses the cross-modal grounding between the text prompt and the image in the cross-attention layer of a diffusion model to segment the most prominent object, referred to by the text prompt in the synthetic image. However, DiffuMasks can only be applied to a generated image. In contrast, our method applies to real images without prompts.


\textbf{Zero-Shot Transfer Segmentation} Another related field is zero-shot segmentation~\cite{kirillov2023segment,shin2022reco,luddecke2022image,bucher2019zero,wu2023diffumask,feng2023network,zhou2022extract}. Works in this line of work possess the capability of ``segmenting anything'' without any training. Most recently, SAM~\cite{kirillov2023segment}, trained on 1.1B segmentation masks, demonstrated impressive zero-shot transfer to any images. Some recent works utilize auxiliary information such as additional images~\cite{shin2022reco,feng2023network} or text inputs~\cite{zhou2022extract,luddecke2022image} to facilitate zero-shot transfer segmentation. In contrast, our method uses a diffusion model to generate segmentation without synthesizing and querying multiple images and, more importantly, without knowing the objects in the image.

\section{Method}
\label{sec:method}

DiffSeg utilizes a pre-trained stable diffusion model~\cite{rombach2022high} and specifically its \textit{self-attention} layers to produce high-quality segmentation masks. We will briefly review the architecture of the stable diffusion model in Sec.~\ref{sec:diffusion_model} and introduce DiffSeg in Sec.~\ref{sec:diffsam}.

\subsection{Stable Diffusion Model Review}
\label{sec:diffusion_model}

The Stable diffusion model~\cite{rombach2022high} is a popular variant of the diffusion model family~\cite{ho2020denoising,saharia2022photorealistic}, a generative model. Diffusion models have a forward and a reverse pass. In the forward pass, at every time step, a small amount of Gaussian noise is iteratively added until an image becomes an isotropic Gaussian noised image. In the reverse pass, the diffusion model is trained to iteratively remove the Gaussian noise to recover the original clean image. The stable diffusion model~\cite{rombach2022high} introduces an encoder-decoder and U-Net design with attention layers~\cite{ho2020denoising}. It first compress an image $x\in\mathbb{R}^{H\times W\times 3}$ into a latent space with smaller spatial dimension $z\in\mathbb{R}^{h\times w\times c}$ using an encoder $z=\mathcal{E}(x)$, which can be de-compressed through a decoder $\Tilde{x}=\mathcal{D}(z)$. All diffusion processes happen in the latent space through a U-Net architecture. The U-Net architecture will be the focus of this paper's investigation.

\looseness=-1 The U-Net architecture~\cite{rombach2022high} consists of a stack of modular blocks (a schematic is provided in Appendix~\ref{sec:architecuture}). Among those blocks, there are 16 specific blocks composed of two major components: a ResNet layer and a Transformer layer. The Transformer layer uses two attention mechanisms: self-attention to learn the global attention across the image and cross-attention to learn attention between the image and optional text input. The component of interest for our investigation is the \textit{self-attention} layer in the Transformer layer. Specifically, there are a total of $16$ self-attention layers distributed in the 16 composite blocks, giving 16 \textit{self-attention tensors}:
\begin{align}
\label{eq:self_attention}
    \mathcal{A}\in\{\mathcal{A}_k\in\mathbb{R}^{h_k\times w_k\times h_k \times w_k}|k=1,...,16\}.
\end{align}
Each attention tensor $\mathcal{A}_k$ is 4-dimensional\footnote{There is a fifth dimension due to the multi-head attention mechanism. Every attention layer has $8$ multi-head outputs. We average the attention tensors along the multi-head axis to reduce to 4 dimensions because they are very similar across this dimension. Please see Appendix~\ref{sec:head_avg} for details.}. Inspired by DiffuMask~\cite{wu2023diffumask}, which demonstrates object grouping in the \textit{cross-attention} layer, where salient objects are given more attention when referred by its corresponding text input, we hypothesize that the unconditional self-attention also contains inherent object grouping information and can be used to produce segmentation masks without text inputs. 

Specifically, for each spatial location $(I,J)$ in the attention tensor, the corresponding 2D \textit{attention map} $\mathcal{A}_k[I,J,:,:]\in\mathbb{R}^{h_k\times w_k}$\footnote{$\sum \mathcal{A}_k[I,J,:,:] = 1$ is a valid probability distribution.} captures the semantic correlation between all locations and the location $(I,J)$. Each location $(I,J)$ corresponds to a region in the original image pixel space, the size of which depends on the receptive field of the tensor.

\begin{figure*}[ht]
     \centering   
         \centering
         \includegraphics[width=0.8\textwidth]{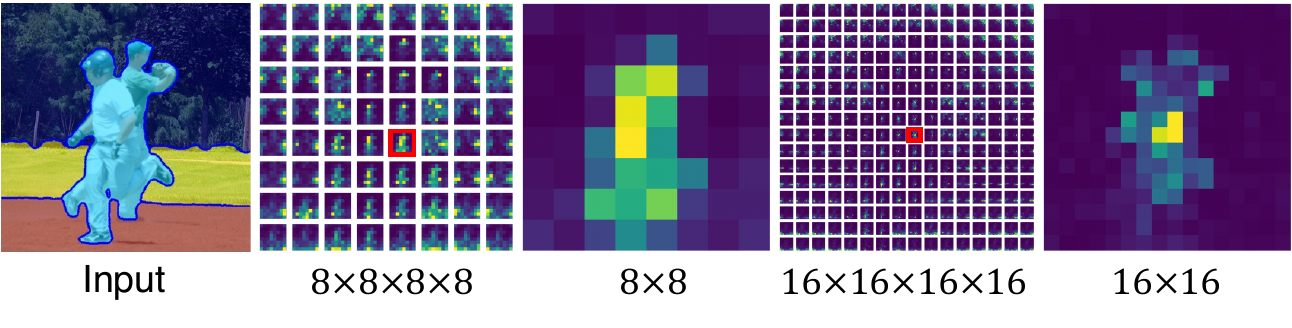}
         \caption{Visualization of Segmentation Masks and Self-Attentions Tensors. Left: Overlay of segmentation and the original image. Right: Attention maps from a stable diffusion model have two properties: Intra-Attention Similarity and Inter-Attention Similarity. Maps of different resolutions have varying receptive fields w.r.t the original image. }
         \label{fig:attention}
\end{figure*}
To illustrate this, we visualize the self-attention tensors $A_7$ and $\mathcal{A}_8$ in Fig.~\ref{fig:attention}. They have a dimension of $\mathcal{A}_7\in\mathbb{R}^{8^4}$ and $\mathcal{A}_8\in\mathbb{R}^{16^4}$\footnote{$\mathbb{R}^{16^4}$ denotes $\mathbb{R}^{16\times16\times16\times16}$.} respectively. Two important observations will motivate our method in Sec.~\ref{sec:diffsam}. 
\begin{itemize}
    \item \textbf{Intra-Attention Similarity}: Within a 2D attention map $\mathcal{A}_k[I,J,:,:]$, locations tend to have strong responses if they correspond to the same  \textit{object group} as $(I,J)$ in the original image space, e.g., $\mathcal{A}_k[I,J,I+1,J+1]$ is likely to be a large value. 
    \item \textbf{Inter-Attention Similarity}: Between two 2D attention maps, e.g., $\mathcal{A}_k[I,J,:,:]$ and  $\mathcal{A}_k[I+1,J+1,:,:]$, they tend to share similar activations if $(I,J)$ and $(I+1,J+1)$ belong to the same \textit{object group} in the original image space. 
\end{itemize}

The second observation is a direct consequence of the first one.  In general, we found that attention maps tend to focus on \textit{object groups}, i.e., groups of individual objects sharing similar visual features (see segmentation examples in Fig.~\ref{fig:domainnet_paint}). In this example (Fig.~\ref{fig:attention}), the two people are grouped as a single \textit{object group}. The $8\times8$ resolution map from a location inside the group highlights much of the entire group. In contrast, the $16\times16$ resolution map from a location inside the object highlights a smaller portion.

Theoretically, the resolution of the attention map dictates the size of its receptive field w.r.t the original image. A smaller resolution leads to a larger respective field, corresponding to a larger portion of the original image. Practically, lower resolution maps, e.g., $8\times8$, provide a better grouping of large objects as a whole, and larger resolution maps, e.g., $16\times16$, give a more fine-grained grouping of components in larger objects and potentially are better for identifying small objects. The current stable diffusion model has attention maps in 4 resolutions $(h_k,w_k)\in\{8\times8,16\times16,32\times32,64\times64\}$. Building on these observations, in the next section, we will propose a simple heuristic to aggregate weights from different resolutions and introduce an iterative method to merge all attention maps into a valid segmentation. 

\textbf{A Note on Extracting Attention Maps.} In our experiments, we use the stable diffusion pre-trained models from Huggingface. These prompt-conditioned diffusion models typically run for $\geq50$ diffusion steps to generate new images. However, in our case, we want to efficiently extract attention maps for an existing clean image without conditional prompts. We achieve this by 1) using only the unconditioned latent and 2) only running the diffusion process once. The unconditional latent is calculated using an unconditioned text embedding. Since we only do one pass through the diffusion model, we also set the time-step variable $t$ to a large value, e.g., $t=300$, in our experiments such that the real images are viewed as primarily denoised generated images from the perspective of the diffusion model.

\begin{figure*}[ht]
     \centering
   
     \begin{subfigure}[b]{0.6\textwidth}
         \centering
         \includegraphics[width=\textwidth]{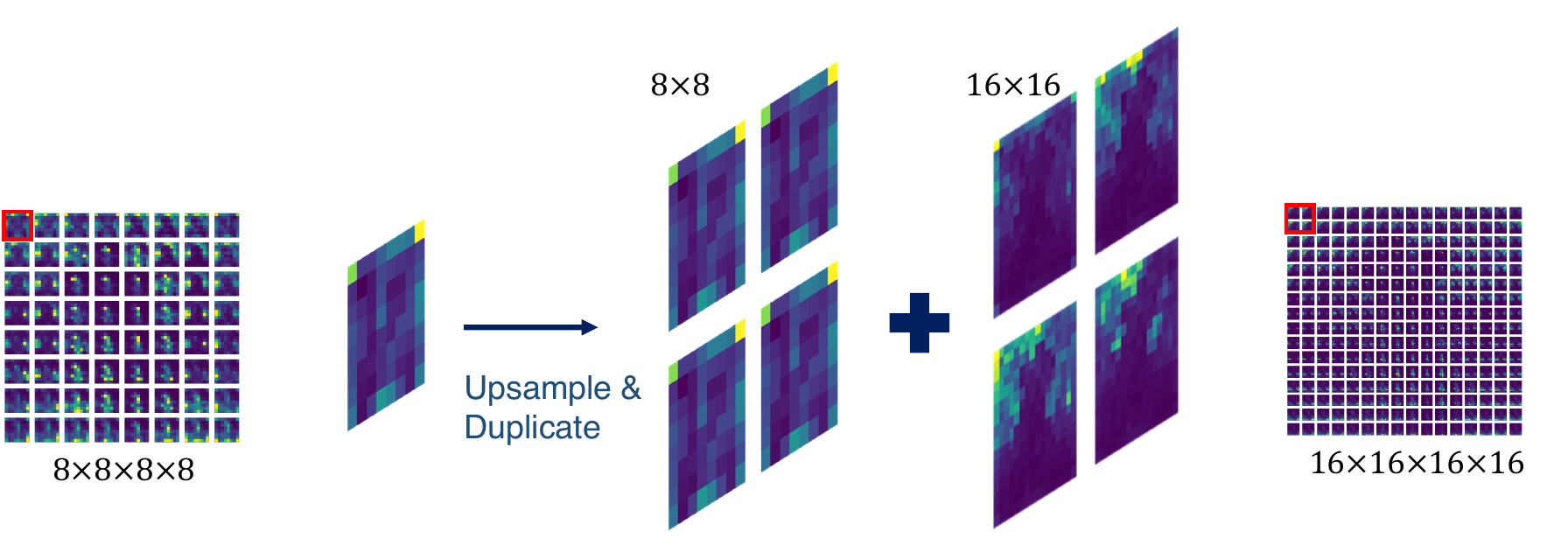}
         \caption{Example of Attention Aggregation. An attention map from a lower resolution is first upsampled and then duplicated to match the receptive field of the higher-resolution maps.}
         \label{fig:aggregation}
     \end{subfigure}
    \hfill
     \begin{subfigure}[b]{0.38\textwidth}
         \centering
         \includegraphics[width=\textwidth]{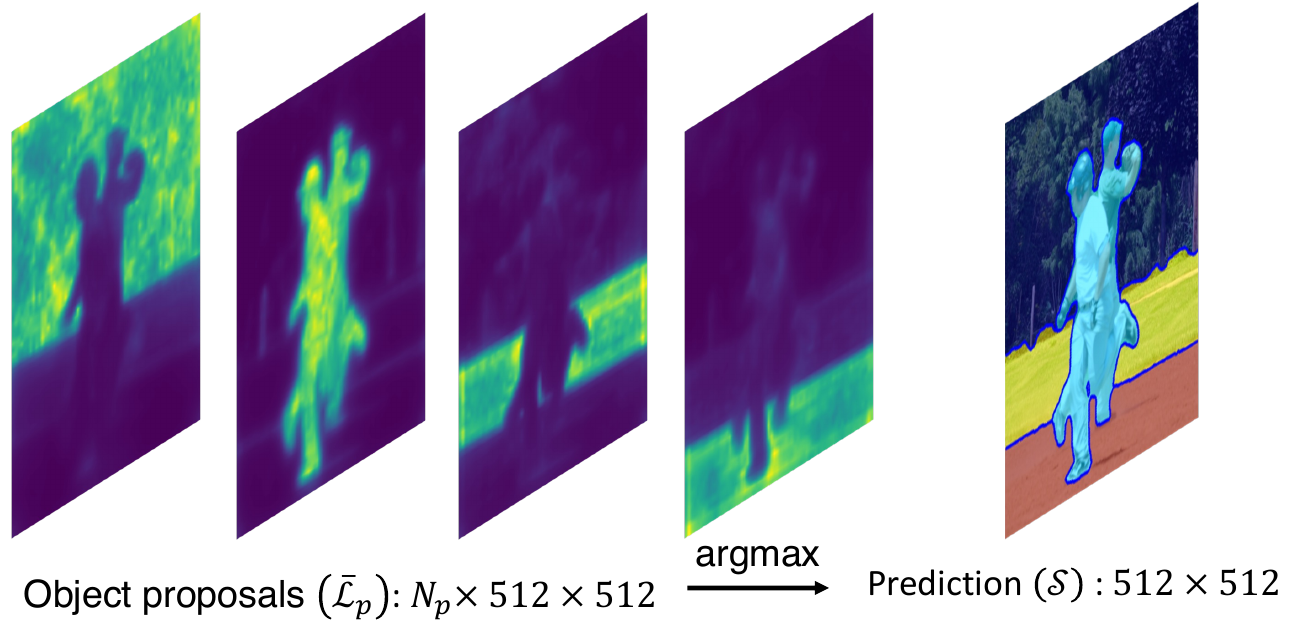}
         \caption{Example of Non-Maximum Suppression (NMS).  NMS looks up the maximum activation across the $\mathcal{L}_p$ proposals for each pixel.}
         \label{fig:nms}
     \end{subfigure}
     \caption{}
\end{figure*}
\subsection{DiffSeg}
\label{sec:diffsam}

Since the self-attention layers capture inherent object grouping information in spatial attention (probability) maps, we propose DiffSeg, a simple post-processing method, to aggregate and merge attention tensors into a valid segmentation mask. The pipeline consists of three components: attention aggregation, iterative attention merging, and non-maximum suppression, as shown in Fig.~\ref{fig:overview}. We will introduce each element in detail next. 

\textbf{Attention Aggregation.}
Given an input image passing through the encoder and U-Net,
a stable diffusion model generates 16 attention tensors. Specifically, there are $5$ tensors for each of the dimensions: $64\times64\times64\times64$, $32\times32\times32\times32$, $16\times16\times16\times16$, $8\times8\times8\times8$; and $1$ tensor for the dimension $8\times8\times8\times8$. The goal is to aggregate attention tensors of different resolutions into the highest resolution tensor. To achieve this, we need to carefully treat the $4$ dimensions differently.

As discussed in the previous section, the 2D map $\mathcal{A}_k[I,J,:,:]$ corresponds to the correlation between all spatial locations and the location $(I,J)$. Therefore, the last $2$ dimensions in the attention tensors are spatially consistent despite different resolutions. Therefore, we upsample (bi-linear interpolation) the last $2$ dimensions of all attention maps to $64\times64$, their highest resolution. Formally, for $\mathcal{A}_k\in\mathbb{R}^{h_k\times w_k, \times h_k \times w_k}$,
\begin{align}
\label{eq:self_upsample}
    \Tilde{\mathcal{A}}_k = \text{Bilinear-upsample}(\mathcal{A}_k)\in \mathbb{R}^{h_k\times w_k \times 64 \times 64}. 
\end{align}

On the other hand, the first $2$ dimensions indicate the locations to which attention maps are referenced. Therefore, we need to aggregate attention maps accordingly. For example, as shown in Fig.~\ref{fig:aggregation}, the attention map in the $(0,0)$ location in $\mathcal{A}_k\in\mathbb{R}^{8^4}$ is first upsampled and then repeatedly aggregated pixel-wise with the $4$ attention maps $(0,0), (0,1), (1,0), (1,1)$ in $\mathcal{A}_z\in\mathbb{R}^{16^4}$. Formally, the final aggregated attention tensor $\mathcal{A}_f\in \mathbb{R}^{64^4}$ is,
\begin{align}
   \mathcal{A}_f[I,J,:,:] &= \sum_{k\in\{1,...,16\}} \Tilde{\mathcal{A}}_k[I/\delta_k,J/\delta_k,:,:]*R_k,\\\nonumber
    &\text{where} \quad \delta_k = 64/w_k,\quad \sum_k R_k = 1.
\end{align}
where $/$ denotes floor division here. Furthermore, to ensure that the aggregated attention map is also a valid distribution, i.e., $\sum \mathcal{A}_f[I,J,:,:] = 1$, we normalize $\mathcal{A}_f[I,J,:,:]$ after aggregation. $R_k$ is the aggregation importance ratio for each attention map. We assign every attention map of different resolutions a weight proportional to its resolution $R_k\propto w_k$. We call this the proportional aggregation scheme (Propto.). The weights $R$ are important hyper-parameters, and we conduct a detailed study of them in Appendix~\ref{sec:addtional_ablation}.

\textbf{Iterative Attention Merging.}
 In the previous step, the algorithm computes an attention tensor $\mathcal{A}_f\in \mathbb{R}^{64^4}$. In this section, the goal is to merge the $64\times64$ attention maps in the tensor $\mathcal{A}_f$ to a stack of \textit{object proposals} where each proposal likely contains the activation of a single object or stuff category.  The naive solution is to run a K-means algorithm on $\mathcal{A}_f$ to find clusters of objects following existing works in unsupervised segmentation~\cite{cho2021picie,hamilton2022unsupervised,li2023acseg}. However, the K-means algorithm requires the specification of the number of clusters. This is not an intuitive hyper-parameter to tune because we aim to segment any image in the wild. Moreover, the K-means algorithm is stochastic depending on the initialization. Each run can have a drastically different result for the same image. To highlight these limitations, we compare them with K-Means baselines in our experiments.
  
 Instead, we propose to generate a sampling grid from which the algorithm can iteratively merge attention maps to create the proposals. Specifically, as shown in Fig.~\ref{fig:overview}, in the \textit{sampling grid generation} step, a set of $M\times M$ ($1\leq M\leq64$) evenly spaced anchor points are generated $\mathcal{M}=\{(i_m,j_m)|m=1,...,M^2\}$. We then sample the corresponding attention maps from the tensor $\mathcal{A}_f$. This operation yields a list of $M^2$ 2D attention maps as anchors,
\begin{align}
    \mathcal{L}_{a} = \{\mathcal{A}_f[i_m,j_m,:,:]\in \mathbb{R}^{64\times64}|(i_m,j_m)\in\mathcal{M}\}.
\end{align}

Since we aim to merge attention maps to find the maps most likely containing an object, we rely on the two observations in Sec.~\ref{sec:diffusion_model}. \textbf{Specifically, when iteratively merging ``similar'' maps, Intra-Attention Similarity \textit{reinforces} the activation of an object and Inter-Attention Similarity \textit{grows} the activation to include as many pixels of the same object within the merged map.} To measure similarity, we use KL divergence to calculate the ``distance" between two attention maps, since each attention map $\mathcal{A}_f[i,j,:,:]$ (abbrev. $\mathcal{A}_f[i,j]$) is a valid probability distribution. Formally,
\begin{align}
    &2*D(\mathcal{A}_f[i,j],\mathcal{A}_f[y,z])\\\nonumber
    &= \left(\mathrm{KL}(\mathcal{A}_f[i,j]\|\mathcal{A}_f[y,z]) + \mathrm{KL}(\mathcal{A}_f[yz,z]\|\mathcal{A}_f[i,j])\right).
\end{align}
We use the forward and reverse KL to address the asymmetry in KL divergence. Intuitively, a small $\mathcal{D}(\cdot)$ indicates a high similarity between two attention maps and that their union is likely to represent better the object they both belong to.

Then, we start the $N$ iterations of the merging process. \textbf{In the first iteration}, we compute the pair-wise distance between each element in the anchor list and all attention maps using $\mathcal{D}(\cdot)$. We introduce a threshold hyper-parameter $\tau$. For each element in the list, we average all attention maps with a distance smaller than the threshold, effectively taking the union of attention maps likely belonging to the same object that the anchor point belongs to. All merged attention maps are stored in a new proposal list $\mathcal{L}_p$.

Note that the first iteration does \textit{not} reduce the number of proposals compared to the number of anchors. \textbf{From the second iteration onward}, the algorithm starts merging maps and reducing the number of proposals simultaneously by computing the distance between an element from the proposal list $\mathcal{L}_p$ and all elements from the \textit{same} list, and merging elements with distance smaller than $\tau$ \textit{without} replacement. The entire iterative attention merging algorithm is provided in Alg.~\ref{alg:merging} (Appendix~\ref{sec:algorithm}).

\textbf{Non-Maximum Suppression.} The iterative attention merging step yields a list $\mathcal{L}_p\in\mathbb{R}^{N_p\times64\times64}$ of $N_p$ object proposals in the form of attention maps (probability maps). Each proposal in the list potentially contains the activation of a single object. To convert the list into a valid segmentation mask, we use non-maximum suppression (NMS). This can be easily done since each element is a probability distribution map. We can take the index of the largest probability at each spatial location across all maps and assign that location's membership to the corresponding map's index.  An example is shown in Fig.~\ref{fig:nms}. Note that, before NMS, we upsample all elements in $\mathcal{L}_p$ to the original resolution.  Formally, the final segmentation mask $\mathcal{S}\in\mathbb{R}^{512\times512}$ is
\begin{align}
\label{eq:diffseg_output}
     \Bar{\mathcal{L}}_p &= \text{Bilinear-upsample}(\mathcal{L}_p) \in \mathbb{R}^{N_p\times512\times512}\\\nonumber
    \mathcal{S}[i,j] &= \argmax \Bar{\mathcal{L}}_p[:,i,j] \quad \forall i,j\in\{0,...,511\}.
\end{align}
\section{Experiments}
\label{sec:experiment}
\textbf{Datasets.} Following existing works in unsupervised segmentation~\cite{ji2019invariant,caron2018deep,cho2021picie,hamilton2022unsupervised,zhou2022extract,shin2022reco}, we use two popular segmentation benchmarks, COCO-Stuff-27~\cite{cordts2016cityscapes,cho2021picie,ji2019invariant} and Cityscapes~\cite{cordts2016cityscapes}. COCO-stuff-27~\cite{cho2021picie,ji2019invariant} is a curated version of the original COCO-stuff dataset~\cite{cordts2016cityscapes}. Specifically, COCO-stuff-27 merges the 80 things and 91 stuff categories in COCO-stuff into 27 mid-level categories. We evaluate our method on the validation split curated by prior works~\cite{cho2021picie,ji2019invariant}. Cityscapes~\cite{cordts2016cityscapes} is a self-driving dataset with 27 categories. We evaluate the official validation split. 
For both datasets, we report results for the diffusion-model-native $512times512$ input resolution and a lower $320times320$ resolution following prior works.

\textbf{Metrics.} Following prior works~\cite{ji2019invariant,caron2018deep,cho2021picie,hamilton2022unsupervised,zhou2022extract,shin2022reco}, we use pixel accuracy (ACC) and mean intersection over union (mIoU) to measure segmentation performance. Because our method does not provide a semantic label, we use the Hungarian matching algorithm~\cite{kuhn1955hungarian} to assign each predicted mask to a ground truth mask. When there are more predicted masks than ground truth masks, the unmatched predicted masks will be considered false negatives. In addition to the metrics, we highlight the requirements other baselines need in inference. Specifically, we emphasize unsupervised adaptation (UA), language dependency (LD), and auxiliary image (AX). UA means that the specific method requires unsupervised training on the target dataset. This is common in the unsupervised segmentation literature. Methods without the UA requirement are considered zero-shot. LD means that the method requires text input, such as a descriptive sentence for the image, to facilitate segmentation. AX means that the method requires an additional pool of reference images or synthetic images.

\textbf{Models.} DiffSeg builds on pre-trained stable diffusion models. We use stable diffusion V1.4~\cite{diffusion_1_4}.

\subsection{Main Results}

\label{sec:main_results}
\begin{table}[t]
    \centering
    \resizebox{.45\textwidth}{!}{
    \begin{tabular}{c|ccc|cc}
    \toprule   
     & \multicolumn{3}{c|}{Requirements} & \multicolumn{2}{c}{Metrics}\\
    \midrule
     Model& LD  & AX & UA & ACC. & mIoU \\
     \midrule
     IIC~\cite{ji2019invariant} & \xmark & \xmark& \cmark & 21.8 & 6.7\\
     MDC~\cite{caron2018deep} & \xmark & \xmark& \cmark & 32.3 & 9.8\\
     PiCLE~\cite{cho2021picie} & \xmark & \xmark& \cmark & 48.1 & 13.8\\
    PiCLE+H~\cite{cho2021picie} & \xmark & \xmark& \cmark & 50.0 & 14.4\\
    STEGO~\cite{hamilton2022unsupervised} & \xmark & \xmark& \cmark & 56.9 & 28.2\\
    ACSeg~\cite{li2023acseg} & \cmark & \xmark& \cmark & - & 28.1\\
    \midrule
    MaskCLIP~\cite{zhou2022extract}& \cmark & \xmark & \xmark& 32.2 & 19.6\\
    ReCo~\cite{shin2022reco}& \cmark & \cmark & \xmark & 46.1 & 26.3\\
    \midrule
    K-Means-C (512)  & \xmark & \xmark& \xmark &  58.9 & 33.7\\
    K-Means-S (512)  & \xmark & \xmark& \xmark &  62.6 & 34.7\\
    DBSCAN (512) & \xmark & \xmark& \xmark &  57.7 & 27.2\\
    \midrule
    Ours: DinoSeg (224) & \xmark & \xmark& \xmark &  68.2 & 39.1\\
    Ours: DiffSeg (320) & \xmark & \xmark& \xmark &  \textbf{72.5} & 43.0\\
    Ours: DiffSeg (512) & \xmark & \xmark& \xmark &  \textbf{72.5} & \textbf{43.6}\\
    \bottomrule
    \end{tabular}}
    \caption{Evaluation on COCO-Stuff-27. Language Dependency (LD), Auxiliary Images (AX), Unsupervised Adaptation (UA). We additionally present baseline results using K-Means. K-Means-C (constant) uses a constant number of clusters, 6. K-means-S (specific) uses a specific number of clusters for each image based on the ground truth. The K-Means results use $512\times512$ resolution.}
    \label{tab:coco_with_k_means}
\end{table}
\begin{table}[t]
    \centering
    \resizebox{.45\textwidth}{!}{
    \begin{tabular}{c|ccc|cc}
    \toprule   
     & \multicolumn{3}{c|}{Requirements} & \multicolumn{2}{c}{Metrics}\\
    \midrule
     Model& LD  & AX & UA & ACC. & mIoU \\
     \midrule
     IIC~\cite{ji2019invariant} & \xmark & \xmark& \cmark & 47.9 & 6.4\\
     MDC~\cite{caron2018deep} & \xmark & \xmark& \cmark & 40.7 & 7.1\\
     PiCLE~\cite{cho2021picie} & \xmark & \xmark& \cmark & 65.5 & 12.3\\
    STEGO~\cite{hamilton2022unsupervised} & \xmark & \xmark& \cmark & 73.2 & 21.0\\
    \midrule
    MaskCLIP~\cite{zhou2022extract}& \cmark & \xmark & \xmark& 35.9 & 10.0\\
    ReCo~\cite{shin2022reco}& \cmark & \cmark & \xmark & 74.6 & 19.3\\
    \midrule
    Ours: DiffSeg (320) & \xmark & \xmark& \xmark & 67.3 & 15.2\\
    Ours: DiffSeg (512) & \xmark & \xmark& \xmark & \textbf{76.0} & \textbf{21.2}\\
    \bottomrule
    \end{tabular}}
    \caption{Evaluation on Cityscapes. Language Dependency (LD), Auxiliary Images (AX), Unsupervised Adaptation (UA).}
    \label{tab:cityscapes}
\end{table}
\begin{figure*}[ht]
     \centering
     \begin{subfigure}[b]{0.19\textwidth}
         \centering
         \includegraphics[width=\textwidth]{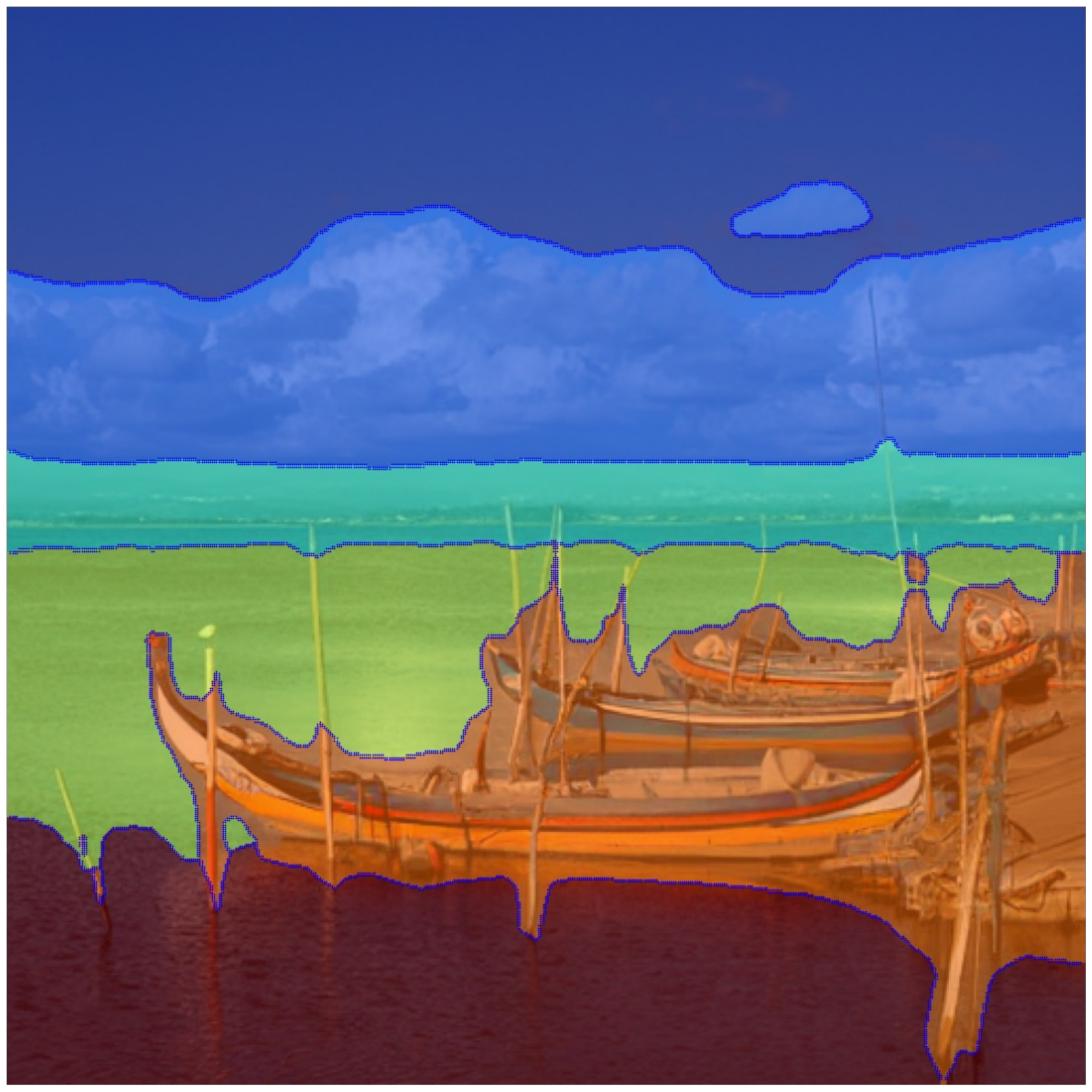}
         \caption{Proportional}
         \label{fig:propto}
     \end{subfigure}
     \begin{subfigure}[b]{0.19\textwidth}
         \centering
         \includegraphics[width=\textwidth]{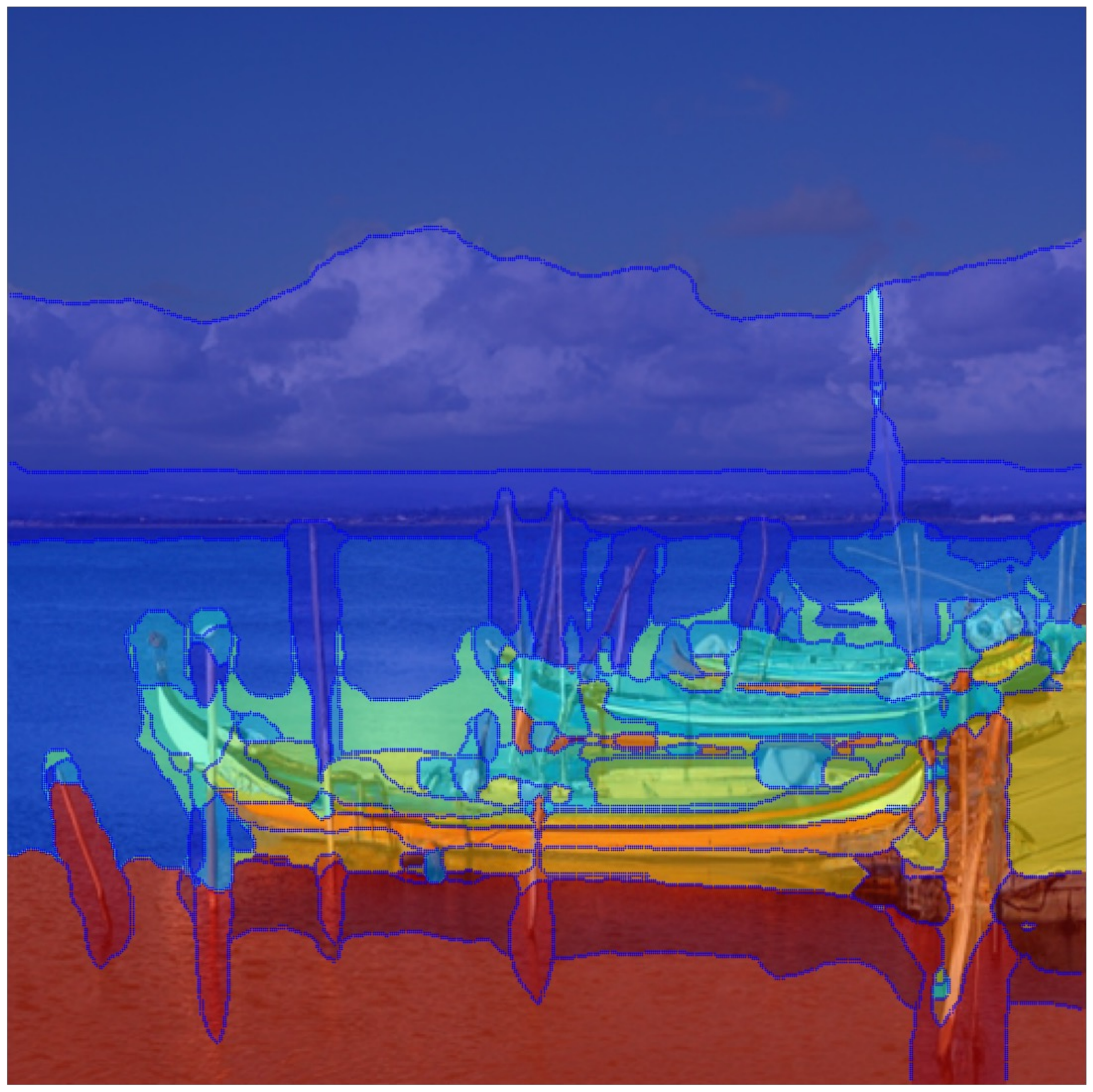}
         \caption{Only $64\times64$ maps}
         \label{fig:only_64}
     \end{subfigure}
     \begin{subfigure}[b]{0.19\textwidth}
         \centering
         \includegraphics[width=\textwidth]{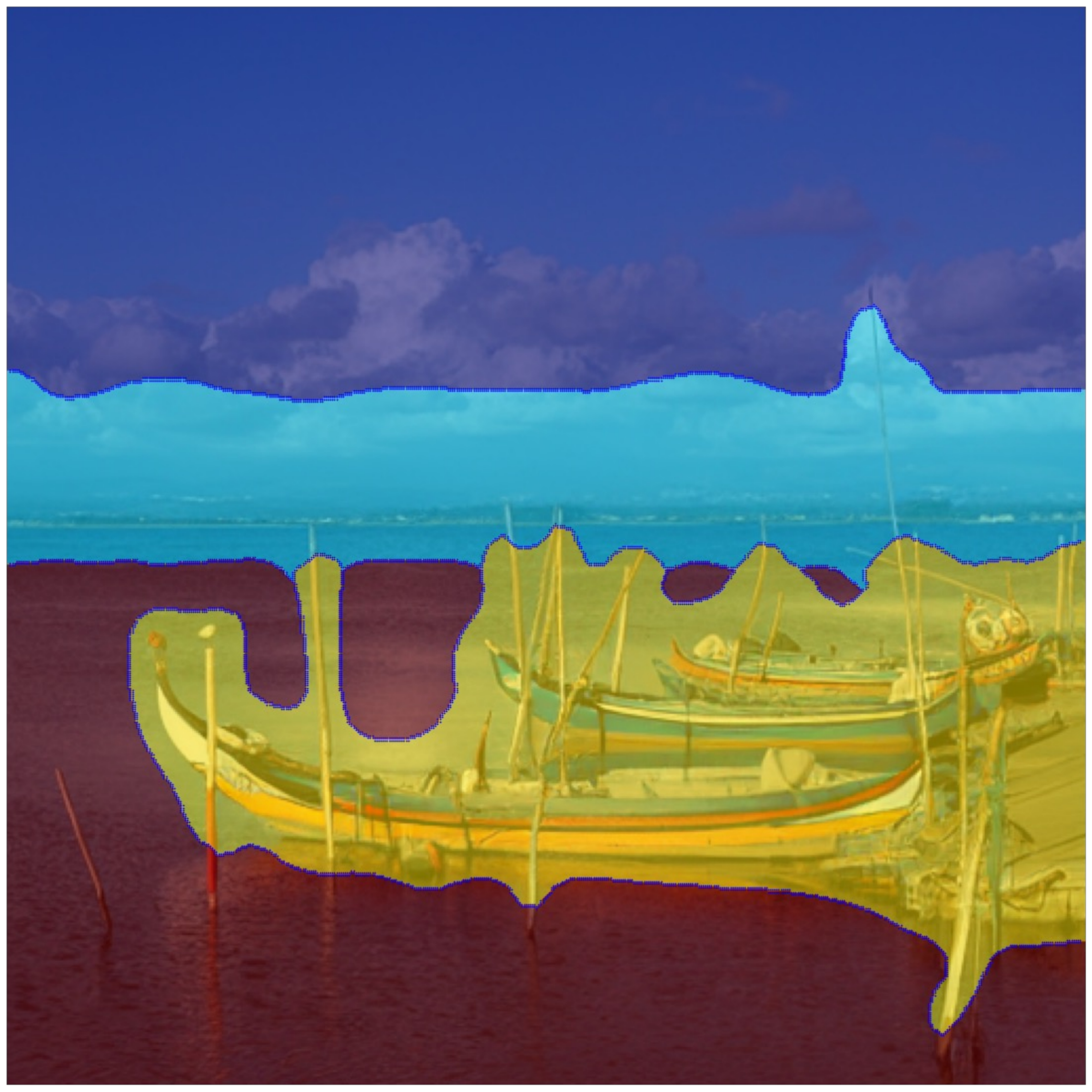}
         \caption{Only $32\times32$ maps}
         \label{fig:only_32}
     \end{subfigure}
     \begin{subfigure}[b]{0.19\textwidth}
         \centering
         \includegraphics[width=\textwidth]{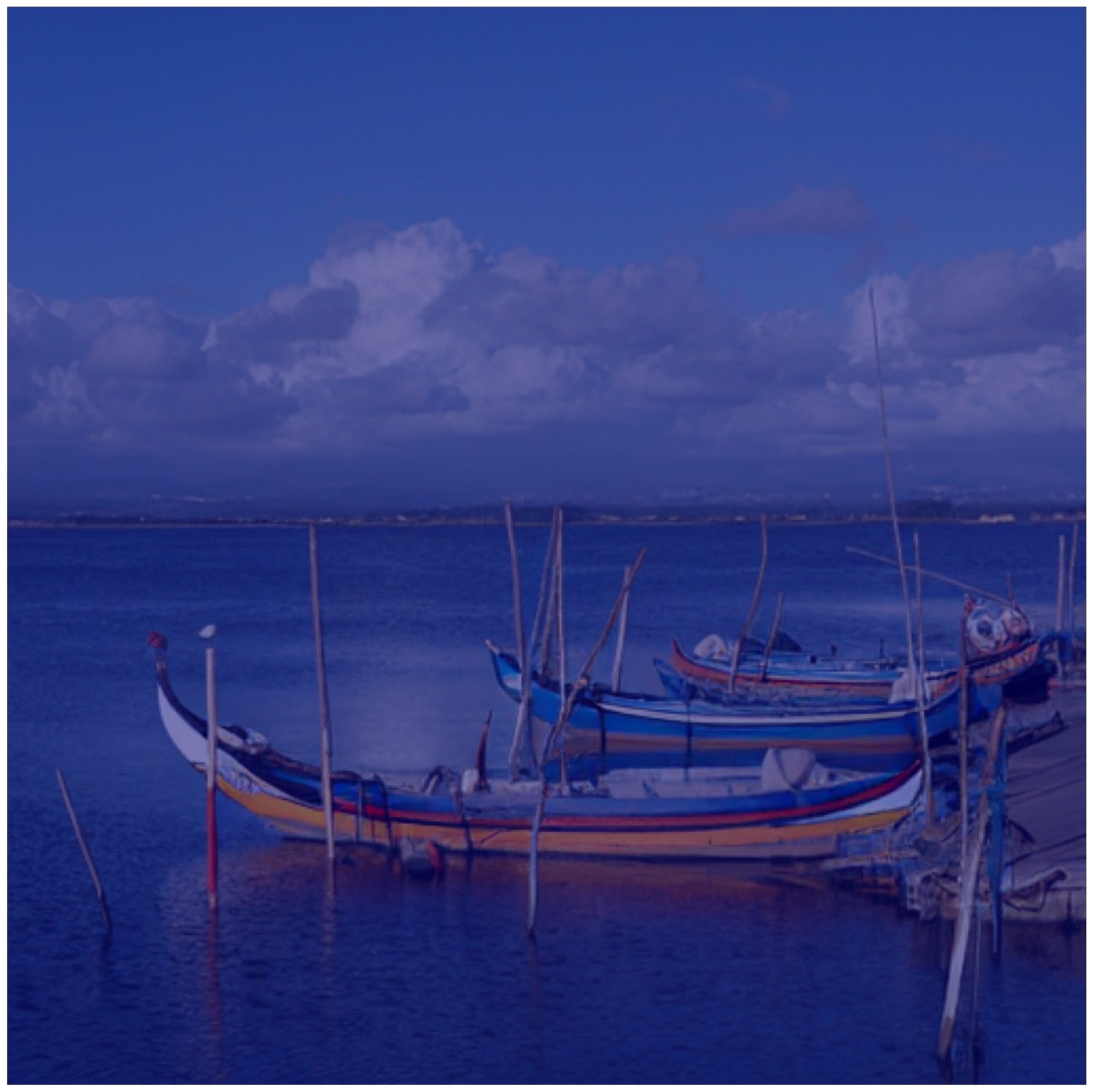}
         \caption{Only $16(8)\times16(8)$ maps}
         \label{fig:only_16}
     \end{subfigure}
     \begin{subfigure}[b]{0.19\textwidth}
         \centering
         \includegraphics[width=\textwidth]{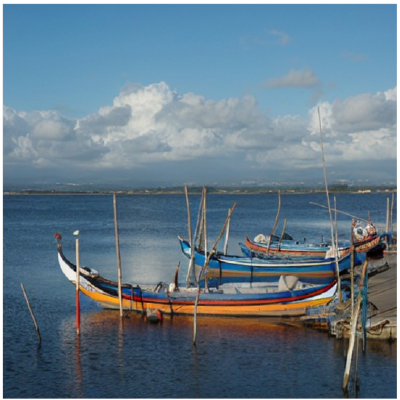}
         \caption{Original image}
         \label{fig:original}
     \end{subfigure}
    \caption{Effects of using Different Aggregation Weights $(R)$. DiffSeg uses a proportional aggregation strategy to balance consistency and detailedness. Higher-resolution maps produce more detailed but fractured segmentation while lower-resolution maps produce more consistent but coarse segmentation. }
    \label{fig:aggregation_weights}
\end{figure*}

\looseness=-1  On the COCO benchmark, we include two K-means baselines, K-Means-C and K-Means-S. K-Means-C uses a constant number of clusters, 6, calculated by averaging over the number of objects in all evaluated images. K-Means-S uses a specific number for each image based on the number of objects in the ground truth of that image. {\color{black} We also include another classic unsupervised clustering algorithm DBSCAN~\cite{ester1996density} that automatically discovers clusters just like DiffSeg.} In Tab.~\ref{tab:coco_with_k_means}, we observe that both K-Means variants outperform prior works, demonstrating the advantage of using self-attention tensors. Furthermore, the K-Means-S variant outperforms the K-Means-C variant. This shows that the number of clusters is an important hyper-parameter and should be tuned for each image. Nevertheless, DiffSeg, despite relying on the same attention tensors $\mathcal{A}_f$, significantly outperforms the K-Means baselines. This proves that our algorithm avoids the disadvantages of K-Means and provides better segmentation. {\color{black} Even though DBSCAN does not require the number of clusters as input, DiffSeg significantly outperforms DBSCAN in our experiments. We hypothesize that this is because DiffSeg encodes spatial prior in its clustering process, i.e., spatially distributed anchors.   
} Please refer to Appendix~\ref{sec:k_means} for more details on implementing the K-Means and DBSCAN baselines. Moreover, DiffSeg significantly outperforms prior works by an absolute $26\%$ in accuracy and $17\%$ in mIoU compared to the prior SOTA zero-shot method, ReCo on COCO-Stuff-27 for both resolutions (320 and 512). 

On the more specialized self-driving segmentation task (Cityscapes), our method is on par with prior works using the smaller 320-resolution input. It outperforms prior works using the larger 512-resolution input in accuracy and mIoU. The resolution of the inputs affects the performance of Cityscapes more severely than that of COCO because Cityscapes has more small classes such as light poles and traffic signs. See discussion on limitations in Appendix~\ref{sec:limitations}. Compared to prior works, DiffSeg achieves this level of performance in a pure zero-shot manner without any language dependency or auxiliary images. Therefore, DiffSeg can segment any image. 

{\color{black}DiffSeg is a generic clustering algorithm, theoretically applicable to any Transformer based model. We further show the performance of a variant (DinoSeg) using DINO~\cite{caron2021emerging} as the backbone in Tab.~\ref{tab:coco_with_k_means}. We observe worse performance due to a smaller attention map size and smaller pre-training dataset. Please refer to Appendix~\ref{sec:dinoseg} for more details and discussion. }

\subsection{Hyper-Parameter Study}
\begin{table}[ht]
    \centering
    \resizebox{.35\textwidth}{!}{
    \begin{tabular}{c|c|c}
    \toprule
          Name & COCO & Cityscapes\\
          \midrule
          Aggregation weights ($R$) & Propto. & Propto.  \\
          Time step ($t$)& 300 & 300\\
          Num. of anchors ($M^2$) & 256 & 256\\
          Num. of merging iterations ($N$) & 3 &3 \\
          KL threshold ($\tau$)& 1.1& 0.9\\
    \bottomrule
    \end{tabular}}
    \caption{Hyper-parameters for Diffseg. The values are used for producing the results in Tab.~\ref{tab:coco_with_k_means} and Tab.~\ref{tab:cityscapes}.}
    \label{tab:hyperparameters}
\end{table}

There are several hyper-parameters in DiffSeg as listed in Tab.~\ref{tab:hyperparameters}. The listed numbers are the exact parameters used in Sec.~\ref{sec:main_results}. This section provides a sensitivity study for these hyper-parameters to illustrate their expected behavior. Specifically, we show the effects of aggregation weights in the main paper and the impact of other hyper-parameters in Appendix~\ref{sec:addtional_ablation}.

\textbf{Aggregation weights $(R)$.}  The first step of DiffSeg is \textit{attention aggregation}, where attention maps of 4 resolutions are aggregated together. We adopt a proportional aggregation scheme. Specifically, the aggregation weight for a map of a certain resolution is proportional to its resolution, i.e., high-resolution maps are assigned higher importance. This is motivated by the observation that high-resolution maps have a smaller receptive field w.r.t the original image, thus giving more details.  To illustrate this, we show the effects of using attention maps of different resolutions for segmentation in Fig.~\ref{fig:aggregation_weights} while keeping other hyper-parameters constant ($t=100, M^2=256, N=3,\tau=1.0$). We observe that high-resolution maps, e.g., $64\times64$ in Fig.~\ref{fig:only_64}, yield the most detailed, however, fractured segmentation. Lower-resolution maps, e.g., $32\times32$ in Fig.~\ref{fig:only_32}, give more coherent segmentation but often over-segment details, especially along the edges. Finally, too low resolutions fail to generate any segmentation in Fig.~\ref{fig:only_16} because the entire image is merged into one object given the current hyper-parameter settings. Notably, our proportional aggregation strategy (Fig.~\ref{fig:propto}) balances consistency and detailedness.

\section{Adding Semantics}
While DiffSeg generates high-quality segmentation masks, like SAM~\cite{kirillov2023segment}, it does not label each mask. We propose a simple extension in Appendix~\ref{sec:semantics} to produce labeled segmentation masks.

\section{Visualization}
\label{sec:vis}
\begin{figure}[ht]
    \centering
    \includegraphics[width=0.45\textwidth]{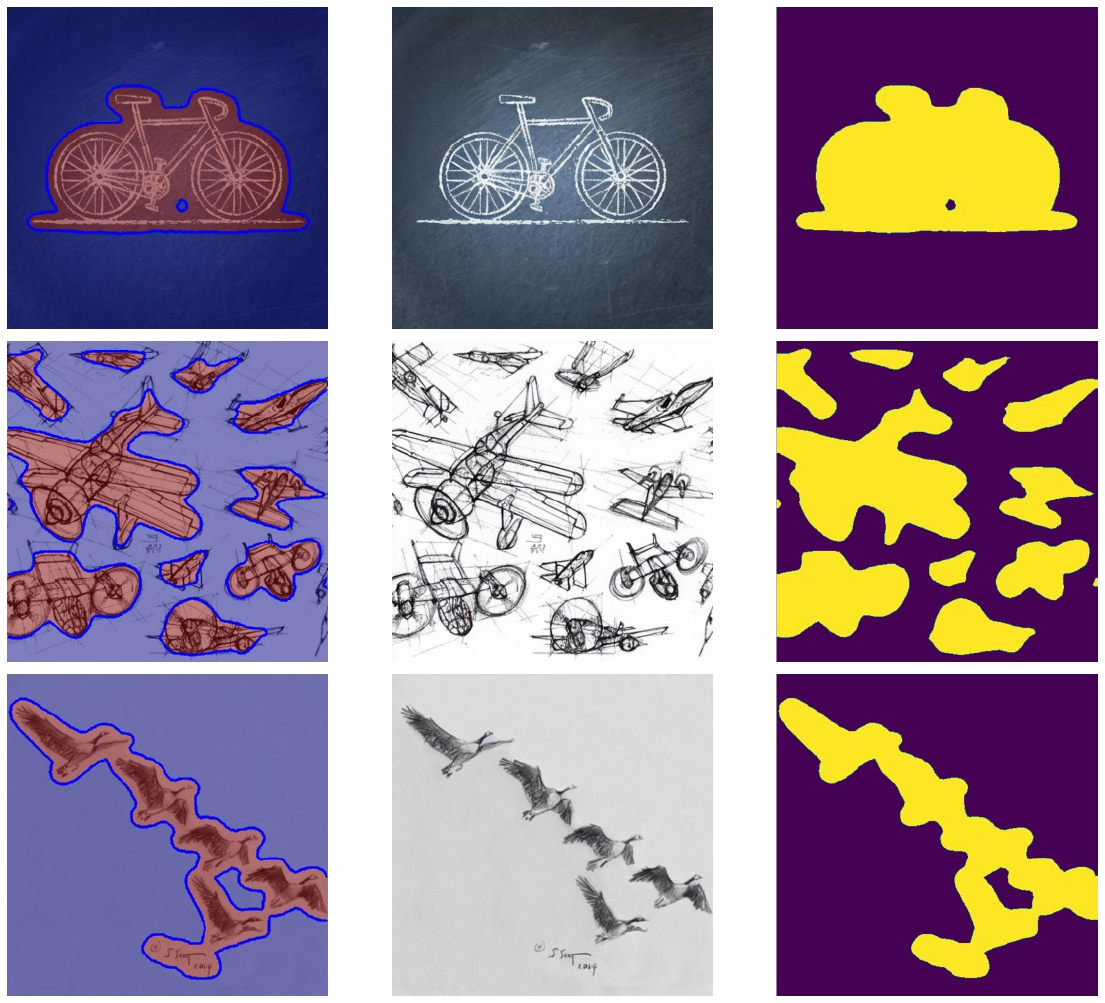}
    \caption{Examples of Segmentation on DomainNet Sketch. Overlay (left), input (middle), and segmentation (right)}
    \label{fig:domainnet_sketch}
\end{figure}
\begin{figure}[ht]
    \centering
    \includegraphics[width=0.45\textwidth]{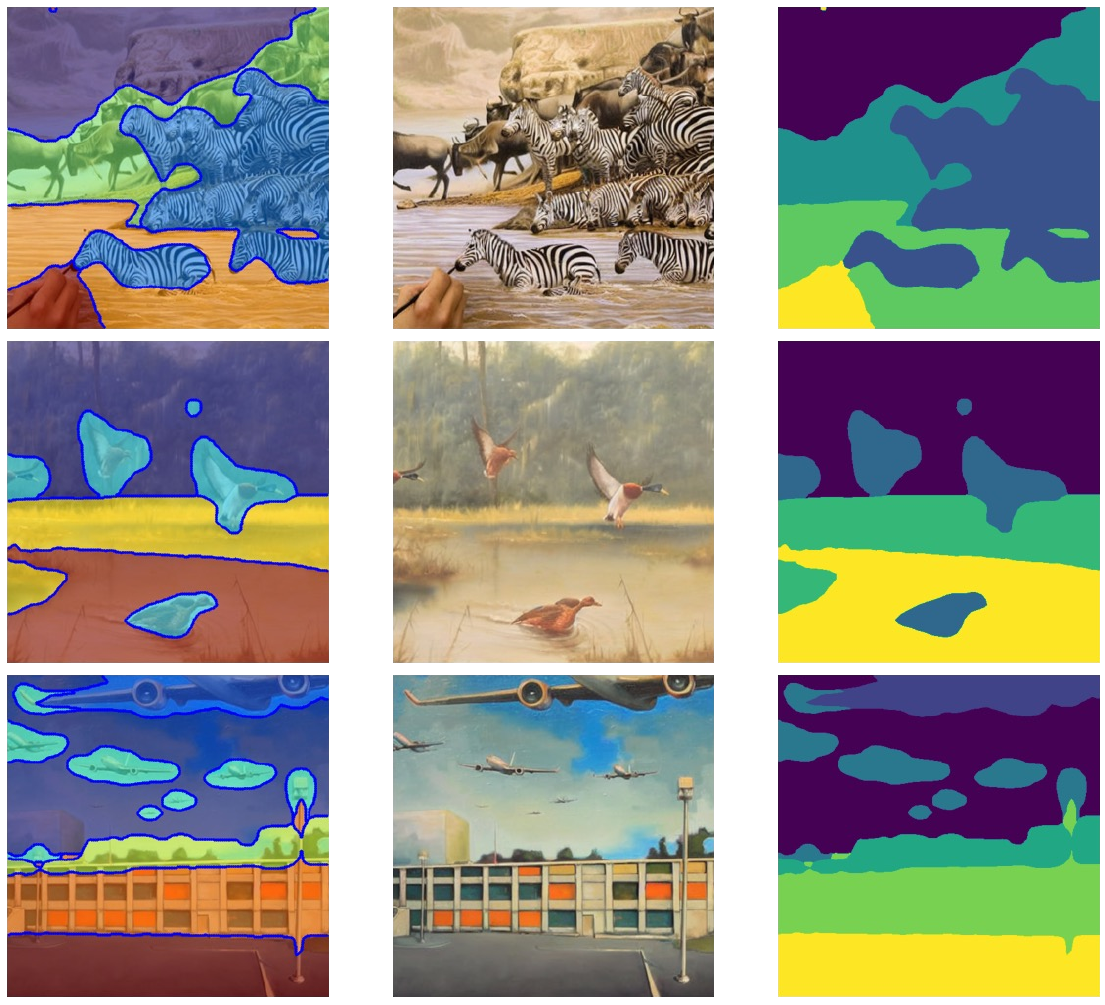}
    \caption{Examples of Segmentation on DomainNet Painting.Overlay (left), input (middle), and segmentation (right)}
    \label{fig:domainnet_paint}
\end{figure}
\begin{figure}[ht]
    \centering
    \includegraphics[width=0.45\textwidth]{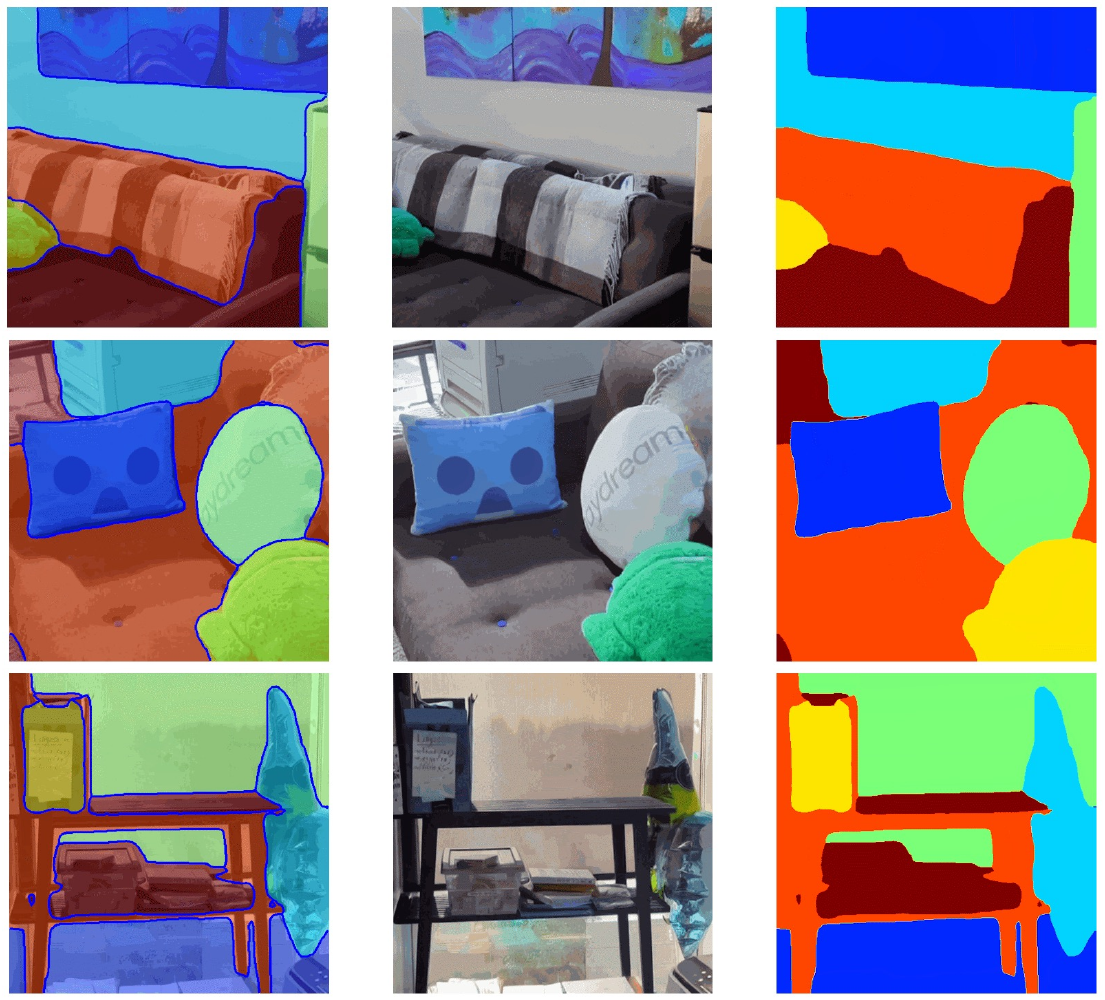}
    \caption{Examples of Segmentation on real-world images captured by a smartphone. Overlay (left), input (middle), and segmentation (right). }
    \label{fig:real_image}
\end{figure}
\begin{figure}[ht]
    \centering
    \vspace{-2mm}
    \includegraphics[width=0.45\textwidth]{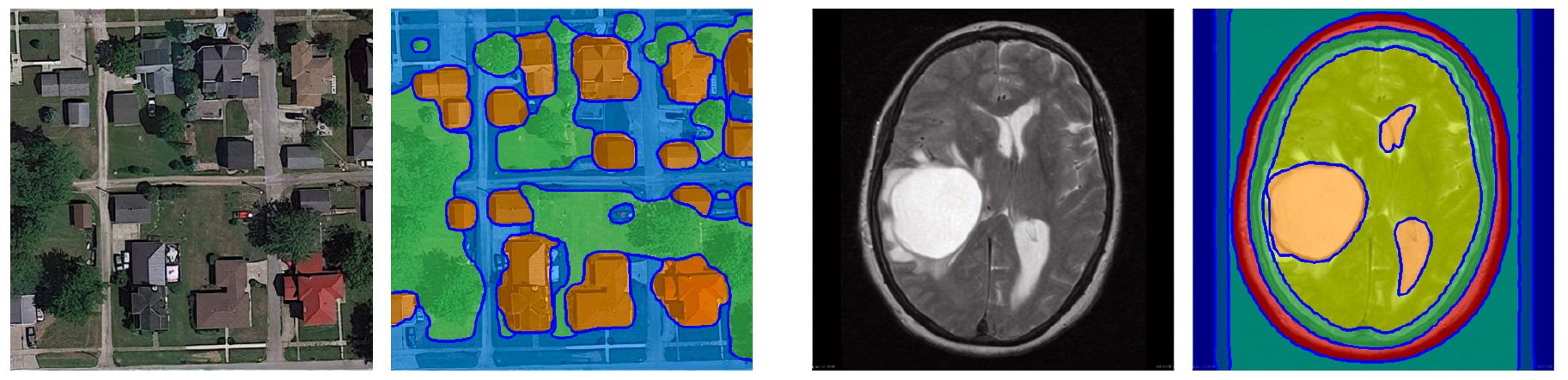}\vspace{-3mm}
    \caption{Segmentation on satellite and brain tumor CT scan.} 
    \label{fig:ood}
    \vspace{-2mm}
\end{figure}
\begin{figure}[ht]
    \centering
    \includegraphics[width=0.45\textwidth]{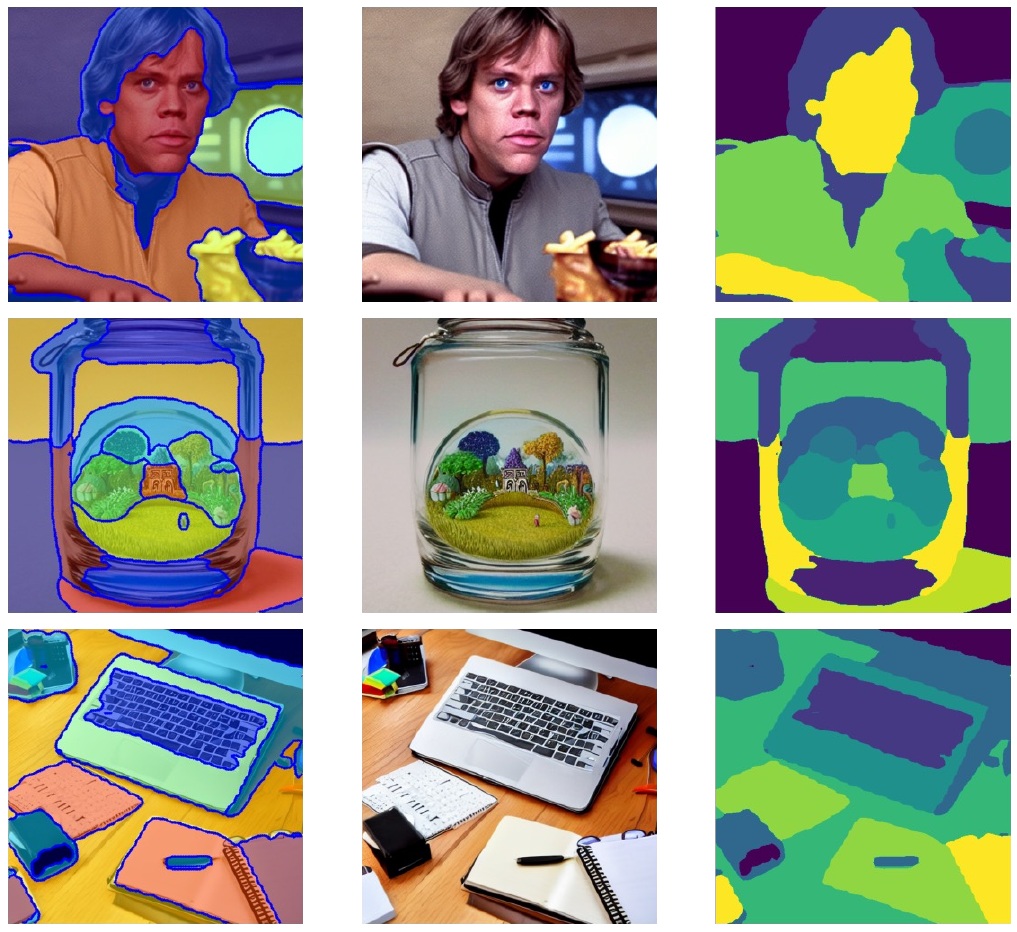}
    \caption{Examples of Segmentation on generated images. Overlay (left), input (middle), and segmentation (right). }
    \label{fig:generated}
\end{figure}

To demonstrate the generalization capability of DiffSeg, we provide examples of segmentation on images of different styles. In Fig.~\ref{fig:domainnet_sketch} and Fig.~\ref{fig:domainnet_paint}, we show segmentation on several sketches and paintings. The images are taken from the DomainNet dataset~\cite{peng2019moment}. In Fig.~\ref{fig:real_image}, we show the segmentation results of real images captured by a smartphone.  We also provide more segmentation examples Cityscapes in Appendix Fig.~\ref{fig:cityscapes_seg} and SUN-RGBD~\cite{zhou2014learning} in Appendix Fig.~\ref{fig:sunrgbd_seg}. {\color{black}In Fig.~\ref{fig:ood}, we show segmentation for satellite images and CT scans.} Naturally, DiffSeg can segment synthetic images of diverse styles, generated by a stable diffusion model.  We show examples in Fig.~\ref{fig:generated} and more in Appendix Fig.~\ref{fig:synthetic_seg}.  

\section{Conclusion}
\label{sec:conclusion}

Unsupervised and zero-shot segmentation is a very challenging setting and only a few papers have attempted to solve it. Most of the existing work either requires unsupervised adaptation (not zero-shot) or external resources. In this paper, we proposed DiffSeg to segment an image without any prior knowledge or external resources, using a pre-trained stable diffusion model, without any additional training. Specifically, the algorithm relies on Intra-Attention Similarity and Inter-Attention Similarity to iteratively merge attention maps into a valid segmentation mask. DiffSeg achieves state-of-the-art performance on popular benchmarks and demonstrates superior generalization to images of diverse styles. 

\section{Acknowledgements}
{\color{black} The work was partially done during the first author's internship at Google. This work was also partially supported by ONR grant N00014-18-1-2829 and Google Gift Grant.}
\newpage
{\small
\bibliographystyle{ieee_fullname}
\bibliography{egbib}
}

\newpage
\section{Appendix}
\label{sec:appendix}
\subsection{Stable Diffusion Architecture}
\label{sec:architecuture}
\begin{figure}[h]
    \centering
    \includegraphics[width=0.5\textwidth]{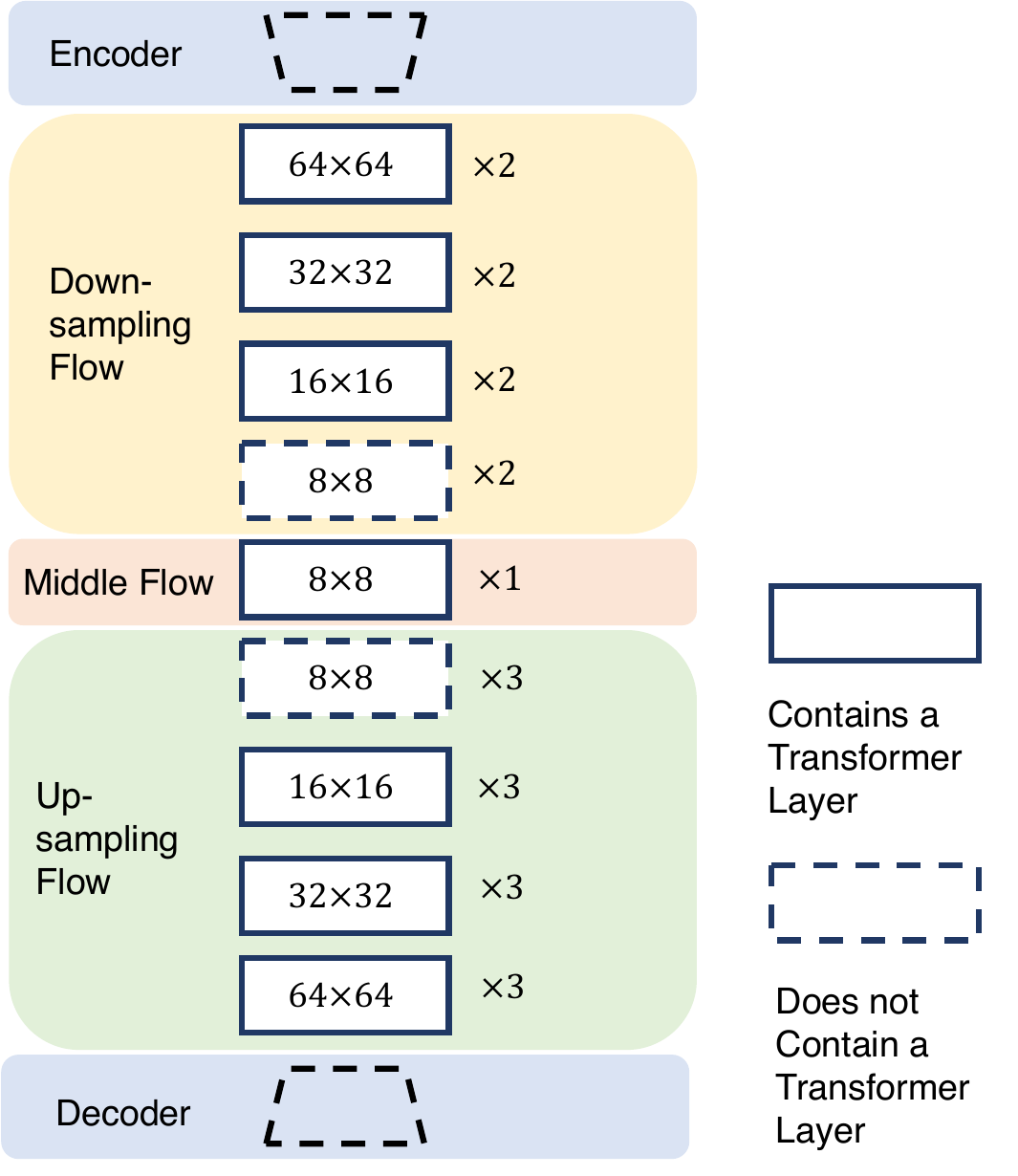}
    \caption{Stable Diffusion Model Schematics. There are in total 16 blocks having Transformer Layers. Each generates a 4D self-attention tensor of different resolutions.}
    \label{fig:schematics}
\end{figure}

\subsection{Discussion on Averaging Multi-head Attentions}
\label{sec:head_avg}
In Sec.~\ref{sec:diffusion_model}, we mentioned that DiffSeg averages the multi-head attention along the multi-head axis, reducing a 5D tensor to 4D. This is motivated by the fact that different attention heads capture very similar correlations. To show this, we calculate the average KL distance across the multi-head axis for all attention tensors of different resolutions used for segmentation. We use 180 training data to calculate the following statistics in Tab.~\ref{tab:head_avg}.  
\begin{table}[ht]
    \centering
    \begin{tabular}{c|c|c}
    \toprule
           & COCO & Cityscapes\\
          \midrule
          Avg. multi-heads KL & 0.46 & 0.45  \\
          Merge Threshold KL ($\tau$)& 1.10 & 0.90\\   
          \bottomrule
    \end{tabular}
    \caption{Average KL distance across the multi-head axis and the merge threshold used in the paper.}
    \label{tab:head_avg}
\end{table}
The difference across the multi-head channels is much smaller than the KL merge threshold used in the paper. This means that the attentions are very similar across the channels and they would have been merged together if not averaged at first because their distance is well below the merge threshold. 

\subsection{Iterative Attention Aggregation}
\label{sec:algorithm}
\begin{algorithm}
\caption{Iterative Attention Merging}\label{alg:merging}
\begin{algorithmic}
\Require $\mathcal{L}_a, \mathcal{A}_f,N,\tau\ $
\State $\mathcal{L}_p = \left\{\frac{1}{|\mathcal{V}|}\sum_{(i,j)\in\mathcal{V}} \mathcal{A}_f[i,j] | v = 1,...,M^2\right\}$
\State $\quad\quad\quad\quad$ where  $\mathcal{V} = \{(i,j)|\mathcal{D}\left(\mathcal{L}_{a}[v],\mathcal{A}_f[i,j]\right)<\tau\}$

\For{$N-1$ iterations}
 \State Initialize $\Tilde{\mathcal{L}}_p = []$
 \For{$\mathcal{A}$ in $\mathcal{L}_p$}
  \State $V=\frac{1}{|\mathcal{V}|}\sum_{v\in\mathcal{V}} \mathcal{L}_p[v]$ \Comment{Merge attention maps}
  \State $\quad\quad\quad$ where  $\mathcal{V}=\{v|\mathcal{D}\left(\mathcal{A},\mathcal{L}_p[v]\right)<\tau\}$
  \State Add $V$ to $\Tilde{\mathcal{L}}_p$
  \State Remove $\mathcal{L}_p[v] \quad \forall v\in\mathcal{V}\quad$ from $\mathcal{L}_p$
 \EndFor
 \State  $\mathcal{L}_p \leftarrow \Tilde{\mathcal{L}}_p$
\EndFor

\end{algorithmic}
\end{algorithm}

\subsection{Comparisons to K-Means and DBSCAN Baselines}
\label{sec:k_means}
DiffSeg uses an iterative merging process to generate the segmentation mask to avoid two major limitations of popular k-means clustering-based algorithms: 1) K-means needs specification of the number of clusters and 2) K-means is stochastic depending on initialization of cluster initialization. In this section, we present a comparison between DiffSeg and a K-means baseline. Specifically, after the \textit{Attention Aggregation} stage, we obtain a 4D attention tensor $\mathcal{A}_f\in\mathbb{R}^{64^4}$. Instead of using iterative merging as in DiffSeg, we directly apply the K-means algorithm on tensor $\mathcal{A}_f$. To do this, we reshape the tensor to $4096\times4096$, which represents 4096 vectors of dimension 4096. The goal is to cluster the 4096 vectors into $N$ clusters. 

We use the Sklearn k-means implementation~\cite{k_means} with k-means$++$ initialization~\cite{arthur2007k}. We present results for the number of clusters using a constant number averaged over all evaluated images and a specific number for each image. The average number and specific number are obtained from the ground truth segmentation masks. We use the COCO-Stuff dataset as the benchmark. 

DBSCAN~\cite{ester1996density} is a classic density-based clustering algorithm that does not require the number of clusters as input. We use the Sklearn DBSCAN implementation~\cite{dbscan} with the default Euclidean metric. We sweep $eps$, which determines the maximum distance for a pair of samples to be considered as neighbors. The same as the K-means implementation, we directly apply DBSCAN to $\mathcal{A}_f$.

\subsection{DinoSeg}
\label{sec:dinoseg}
DiffSeg is a generic unsupervised clustering algorithm theoretically applicable to any Transfer features backbone. In this section, we adopt DiffSeg to a DINO backbone~\cite{caron2021emerging}. Specifically, we use a DINO-Base backbone with a patch size of $8$. DINO takes images at a resolution of $224$. 

The pipeline is similar to that of DiffSeg, the only difference being the attention aggregation. DINO produces attention tensors of the \textit{same} spatial size $28\times28$ for all its layers. Therefore, aggregation simply averages over all 12 layers and 12 multi-head channels. 

As shown in Tab.~\ref{tab:coco_with_k_means}, the performance of DinoSeg is worse than that of DiffSeg. There can be multiple reasons. 1) DINO is pre-trained on ImageNet1K, which is much smaller than LAION5B~\cite{schuhmann2022laion}. 2) DINO's input images are of lower resolution. 3) The spatial dimension of DINO's attention outputs is only $28\times28$, which is smaller than the $64\times$ resolution found in Stable Diffusion.

\subsection{Additional Ablation study}
\label{sec:addtional_ablation}
\begin{figure*}[ht]
     \centering
     \begin{subfigure}[b]{0.24\textwidth}
         \centering
         \includegraphics[width=\textwidth]{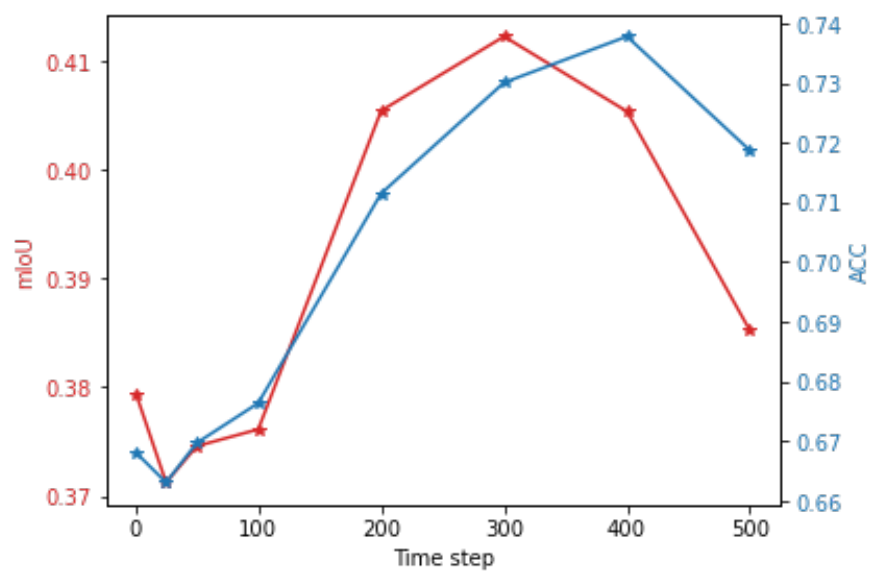}
         \caption{Time Step $t$}
         \label{fig:time_step}
     \end{subfigure}
     \begin{subfigure}[b]{0.24\textwidth}
         \centering
         \includegraphics[width=\textwidth]{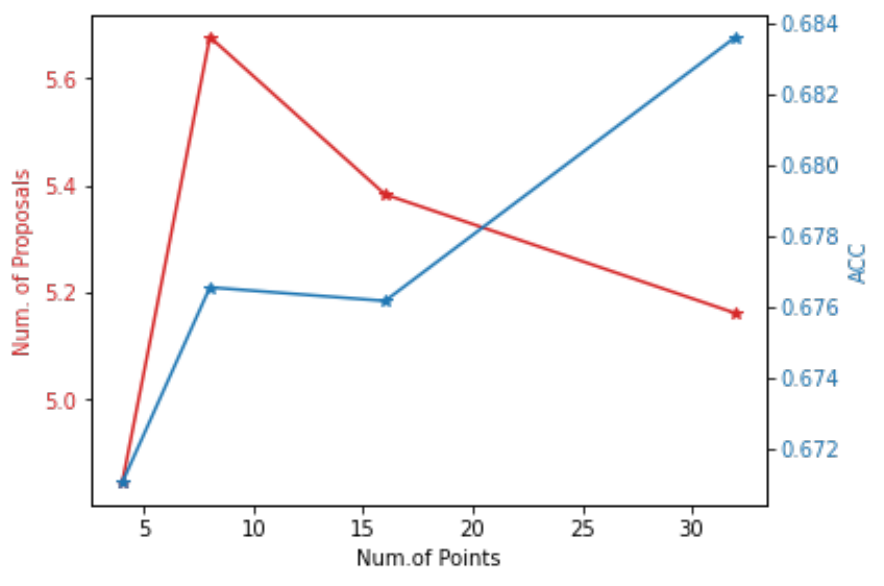}
         \caption{Number of Anchors $M^2$}
         \label{fig:num_anchors}
     \end{subfigure}
     \begin{subfigure}[b]{0.24\textwidth}
         \centering
         \includegraphics[width=\textwidth]{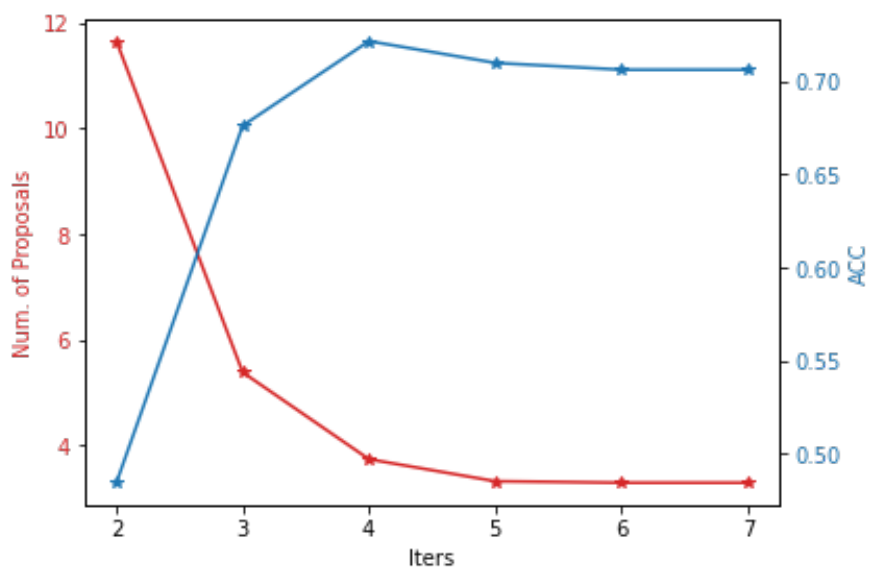}
         \caption{Number of Merging Iters. ($N$)}
         \label{fig:num_merging_iters}
     \end{subfigure}
     \begin{subfigure}[b]{0.24\textwidth}
         \centering
         \includegraphics[width=\textwidth]{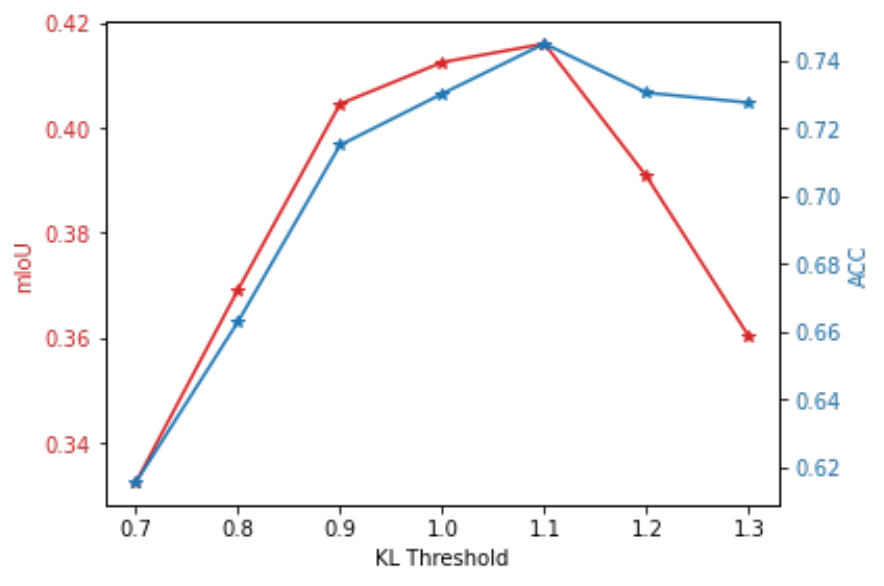}
         \caption{KL threshold $\tau$}
         \label{fig:kl_threshold}
     \end{subfigure}
    \caption{Effects of the Time Step $t$, the Number of Anchors $M^2$, the Number of Merging Iterations ($N$) and the KL Threshold $\tau$ on COCO-Suff-27. The same set of $(R,t,M,N)$ works generally well for different settings (Tab.~\ref{tab:hyperparameters}). A reasonable range for $\tau$ is $0.9\sim1.1$.}
    \label{fig:other_para}
\end{figure*}

Most of them have a reasonable range that works well for general settings. Therefore, we do not tune them for each dataset and model. One exception is the KL threshold parameter $\tau$. It is the most sensitive parameter as it directly controls the attention-merging process. We tune this parameter in our experiments on a small subset ($180$ images) from the training set from the respective datasets. All sensitivity study experiments are also conducted on the respective subsets.

\textbf{Time step $(t)$.} The stable diffusion model requires a time step $t$ to indicate the current stage of the diffusion process. Because DiffSeg only runs a single pass through the diffusion process, the time step becomes a hyper-parameter. In Fig.~\ref{fig:time_step}, we demonstrate the effects of setting this parameter to different numbers $t\in\{1,100,200,300,400,500\}$ while keeping the other hyper-parameters constant ($R=\text{propto.}, M^2=256, N=3,\tau=1.0$). In the figure, we observe a general upward trend for accuracy and mIoU when increasing the time step, which peaked around $t=300$. Therefore, we use $t=300$ for our main results.

\textbf{Number of anchors $(M^2)$.}
DiffSeg generates a sampling grid of $M^2$ anchors to start off the attention-merging process. In Fig.~\ref{fig:num_anchors}, we show the number of proposals and accuracy with different numbers of anchor points $M\in\{4,8,16,32\}$ while keeping the other hyper-parameters constant ($R=\text{propto.}, t=100, N=3,\tau=1.0$). We observe that the number of anchor points does not significantly affect the performance of COCO-Stuff-27. Therefore, we keep the default $M=16$.

\textbf{Number of merging iterations $(N)$.} 
The iterative attention merging process runs for $N$ iterations. Intuitively, the more iterations, the more proposals will be merged. In Fig.~\ref{fig:num_merging_iters}, we show the effects of increasing the number of iterations $N\in\{2,3,4,5,6,7\}$ in terms of the number of final objects and accuracy while keeping the other hyper-parameters constant ($R=\text{propto.}, M^2=256, t=100, \tau=1.0$). We observe that at the 3rd iteration, the number of proposals drops to a reasonable amount and the accuracy remains similar afterward. Therefore, we use $K=3$ for a better system latency and performance trade-off. 

\textbf{KL threshold $(\tau)$.} 
The iterative attention merging process also requires specifying the KL threshold $\tau$. It is arguably the most sensitive hyper-parameter and should be tuned preferably separately for each dataset. Too small a threshold leads to too many proposals and too large leads to too few proposals. In Fig.~\ref{fig:kl_threshold}, we show the effect of $\tau\in\{0.7,0.8,0.9,1.0,1.1,1.2,1.3\}$ while using the validated values for the other hyper-parameters ($R=\text{propto.}, M^2=256, t=100, N=3$). We observe that a range $\tau\in[0.9,1.1]$ should yield reasonable performance. We select $\tau=1.1$ for COCO-Stuff-27 and $\tau=0.9$ for Cityscapes, identified using the same procedure. 

\textbf{A Note on the Hyper-Parameters.} We found that the same set of $(R,t,M,N)$ works generally well for different settings. The only more sensitive parameter is $\tau$. A reasonable range for $\tau$ is between 0.9 and 1.1. We would suggest using the default $\tau=1.0$ for the segmentation of images in the wild. For the best benchmark results, proper hyper-parameter selection is preferred. 

\subsection{Limitations}
\label{sec:limitations}
While the zero-shot capability of DiffSeg enables it to segment almost any image, thanks to the generalization capability of the stable diffusion backbone, its performance on more specialized datasets such as self-driving datasets, e.g., Cityscapes, is far from satisfaction. There are several potential reasons. 1) The resolution of the largest attention map is $64\times64$, which can be too small to segment small objects in a self-driving dataset. 2) Stable diffusion models have limited exposure to self-driving scenes. The performance of zero-shot methods largely relies on the generalization capability of the pre-trained model. If during the pre-training stage, the stable diffusion model has not been exposed to a vehicle-centric self-driving scene, it could negatively affect the downstream performance. 3) While DiffSeg is much simpler than competing methods such as ReCo~\cite{shin2022reco}, which requires image retrieval and co-segmentation of a batch of images, it is still not a real-time algorithm. This is because the current stable diffusion model is very large, even though DiffSeg only runs the diffusion step once.  Also, the attention aggregation and merging process are iterative, which incurs higher computation costs for CPU and GPU. 

\subsection{Adding Semantics}
\begin{figure*}[ht]
    \centering
    \includegraphics[width=1.0\textwidth]{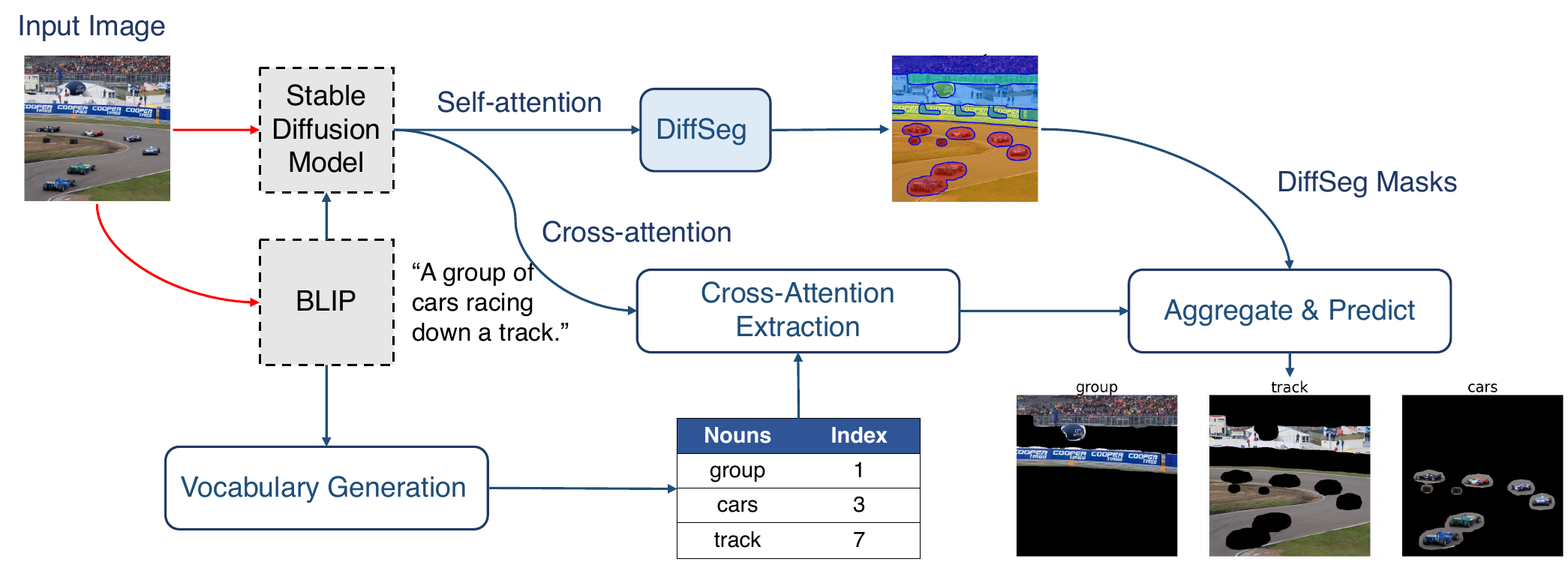}
    \caption{Overview of Semantic DiffSeg.  Semantic DiffSeg extends DiffSeg to add labels to generated masks. It has three major additional components Vocabulary Generation, Cross-Attention Extraction and Aggregate $\&$ Predict.}
    \label{fig:semantic_diffseg}
\end{figure*}
\label{sec:semantics}
While DiffSeg generates high-quality segmentation masks, like SAM~\cite{kirillov2023segment}, it does not provide labels for each mask. Inspired by recent open-vocabulary semantic segmentation methods~\cite{xu2023open,liang2023open}, which use an off-the-shelf image captioning model to generate vocabularies for an image and DiffuMask~\cite{wu2023diffumask}, which demonstrates the grounding capability of the \textit{cross-attention} layers in a diffusion model, we propose a simple extension to DiffSeg to produce labeled segmentation masks. We refer to the extended version as Semantic DiffSeg. Specifically, Semantic DiffSeg has three additional components: vocabulary generation, cross-attention extraction, and aggregation $\&$ prediction as shown in Fig.~\ref{fig:semantic_diffseg}.

\textbf{Vocabulary Generation.} Following existing works in open-vocabulary~\cite{xu2023open,liang2023open} segmentation, we use an off-the-shelf image captioning model, e.g., BLIP~\cite{li2022blip} to generate a caption for a given image. Denote $\mathcal{W} = \{w_1,w_2,...,w_K\}$ as the caption where each $w$ represents a word in the caption. To construct the output label space, we extract all \textit{nouns} from the caption using the NLTK toolkit~\cite{loper2002nltk}. Importantly, we also keep track of the \textit{relative position }of the noun word in the caption for indexing purposes. Formally, let $\mathcal{N} = \{n_1,n_2,...n_L\}$ denote the set of extracted nouns from $\mathcal{W}$ and $\mathcal{I} = \{i_1,i_2,...,i_L\}$ denote the set of indices of the relative position corresponding to each noun in the original caption $\mathcal{W}$. 

\textbf{Cross-Attention Extraction.}
Unlike the vanilla DiffSeg, which uses the \textit{unconditional} generation capability of stable diffusion models for segmentation without a prompt input, Semantic DiffSeg adds the generated caption $\mathcal{W}$ as another input to extract meaningful grounded cross-attention maps. DiffuMask~\cite{wu2023diffumask} shows that cross-attention maps corresponding to noun tokens provide grounding for their respective concepts. Inspired by this finding, we also extract the corresponding cross-attention maps from different resolutions using the indices $\mathcal{I}$. Similar to the self-attention formulation in Eq.~\ref{eq:self_attention}, there are a total 16-cross attention layers in a stable diffusion model, giving 16 \textit{cross-attention tensor}:
\begin{align}
\label{eq:cross_attention}
    \mathcal{A_C}\in\{\mathcal{A}_c\in\mathbb{R}^{h_c\times w_c \times Q}|c=1,...,16\}.
\end{align}

where $Q$ is the number of tokenized words in the input caption sequence, e.g., $Q=77$ for stable diffusion models. Unlike the self-attention tensor, the cross-attention tensor is a 3-dimensional tensor, where each  $\mathcal{A}_c[:,:,q]\in\mathbb{R}^{h_c\times w_c}, \forall q\in\{1,..,Q\}$ is the \textit{un-normalized} attention map w.r.t the token $q$. We now extract cross-attention maps corresponding to only the nouns in our caption using our index set $\mathcal{I}$. Note that the caption length $K$ is usually smaller than the token sequence length $Q$. Stable diffusion models use BOS and EOS paddings to unify all input lengths to $Q$. Therefore, in our case, the correct index of a noun word w.r.t the padded tokens is its index $i+1$ resulting from the BOS token offset. Finally, we obtain a cross-attention tensor $\mathcal{A_N}$ corresponding only to the noun tokens.
\begin{align}
\label{eq:extracted_cross_attention}
    \mathcal{A_N}\in\{\mathcal{A}_n\in\mathbb{R}^{h_n\times w_n \times L}|n=1,...,16\}.
\end{align}

\textbf{Aggregation and Prediction.} Similar to the self-attention maps, the cross-attention maps in $\mathcal{A_N}$ are of different resolutions. To aggregate them, we upsample the \textit{first} 2 dimensions of each map to $512$. 
\begin{align}
\label{eq:cross_upsample}
    \Tilde{\mathcal{A}}_n = \text{Bilinear-upsample}(\mathcal{A}_n)\in \mathbb{R}^{ 512 \times 512 \times L}. 
\end{align}
Note this is slightly different from the upsampling of the self-attention maps in Eq.~\ref{eq:self_upsample}, which upsamples the \textit{last} 2 dimensions. Finally, we \textit{sum and normalize} all cross-attention maps to obtain the aggregated cross-attention tensor $\mathcal{A}_{nf}\in \mathbb{R}^{ 512 \times 512 \times L}$. Specifically, the aggregated cross-attention map $\mathcal{A}_{nf}^l \in \mathbb{R}^{ 512 \times 512}$ corresponding to token $l$ is
\begin{align}
\label{eq:self_upsample}
    \mathcal{A}_{nf}^l = \frac{\sum_{n=1}^{16} \Tilde{\mathcal{A}}_n[:,:,l]}{\sum_{w=1}^{512}\sum_{h=1}^{512}\sum_{n=1}^{16} \Tilde{\mathcal{A}}_n[w,h,l]} .
\end{align}

To obtain the final labeled masks, we combine $ \mathcal{A}_{nf}$ with the output from DiffSeg $\Bar{\mathcal{L}}_p \in \mathbb{R}^{N_p\times512\times512}$ in Eq.~\ref{eq:diffseg_output}, where $N_p$ denotes the number of generated masks and each $\Bar{\mathcal{L}}_p[n_p,:,:] \in \mathbb{R}^{512\times512}$ is a binary mask. Specifically, for each mask $ \Bar{\mathcal{L}}_p[n_p,:,:], n_p \in \{1,...,N_p\}$, we calculate the prediction vector $l_{n_p}\in\mathbb{R}^{L}$. Each element  $l_{n_p}^i$ in $l_{n_p}$ is calculated as
\begin{align}
     l_{n_p}^i=  \Bar{\mathcal{L}}_p[n_p,:,:] \circledast \mathcal{A}_{nf}[:,:,i],
\end{align}
where $\circledast$ denotes the element-wise product. Finally, the label for mask $n_p$ is obtained by taking the maximum element in the prediction vector $l_{n_p}$. As a post-processing step, we merge all masks with the same label into a single mask. 

\begin{figure}[ht]
    \centering
    \includegraphics[width=0.45\textwidth]{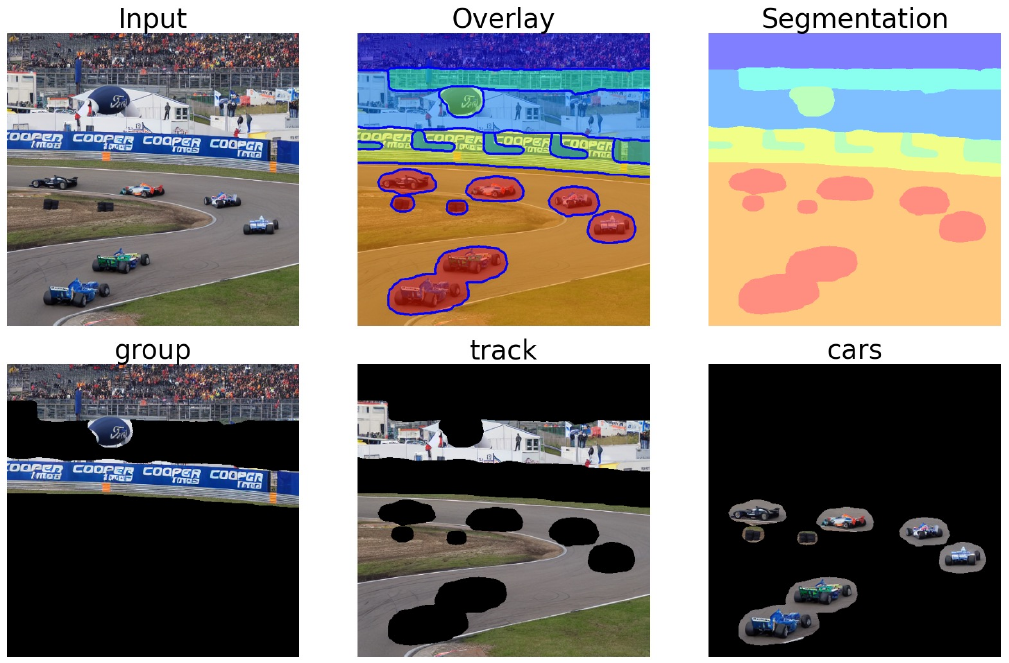}
    \caption{Semantic DiffSeg Example. First row: DiffSeg output. Second row: semantic segmentation with mask merging. Generated caption: A group of cars racing down a track.}
    \label{fig:semantics1}
\end{figure}
\begin{figure}[ht]
    \centering
    \includegraphics[width=0.45\textwidth]{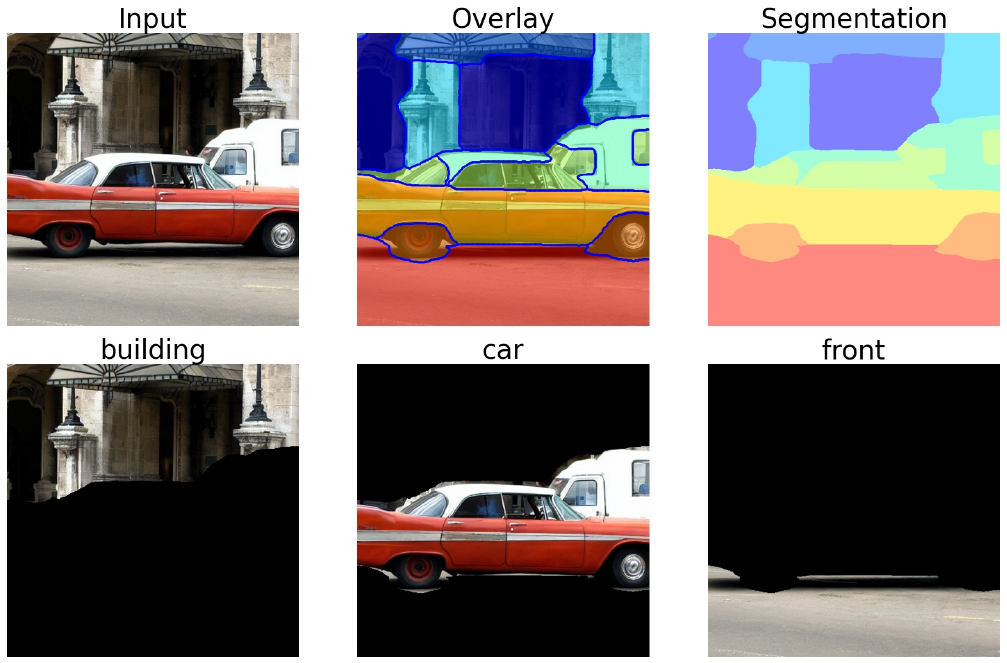}
    \caption{Semantic DiffSeg Example. First row: DiffSeg output. Second row: semantic segmentation with mask merging. Generated caption: A red car parked in front of a building.}
    \label{fig:semantics1}
\end{figure}

\clearpage
\newpage
\subsection{Additional Visualization}
\begin{figure}[ht]
    \centering
    \includegraphics[width=0.5\textwidth]{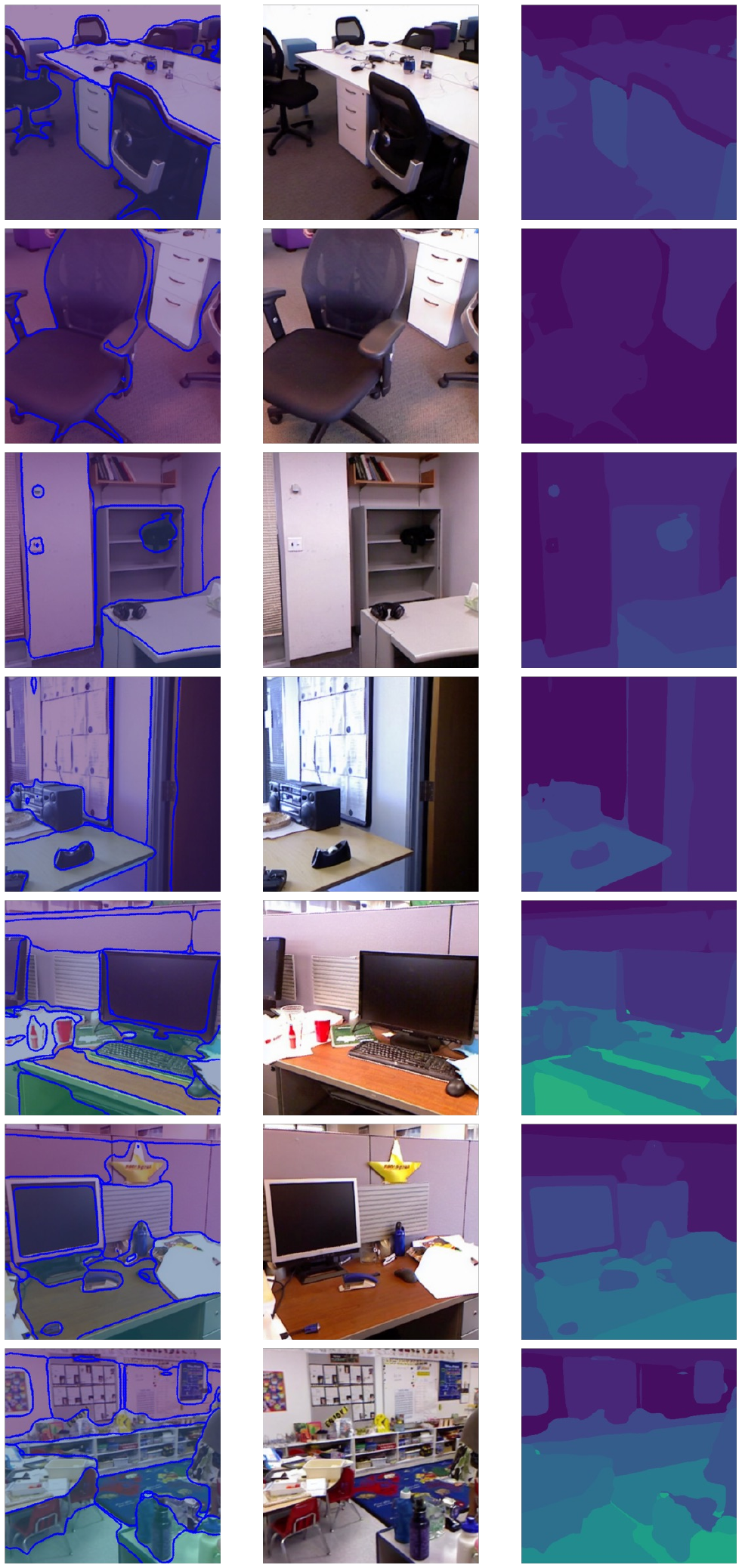}
    \caption{Examples of Segmentation on SUN-RGBD Images. Overlay (left), Input (middle), and segmentation (right)}
    \label{fig:sunrgbd_seg}
\end{figure}
\begin{figure}[ht]
    \centering
    \includegraphics[width=0.5\textwidth]{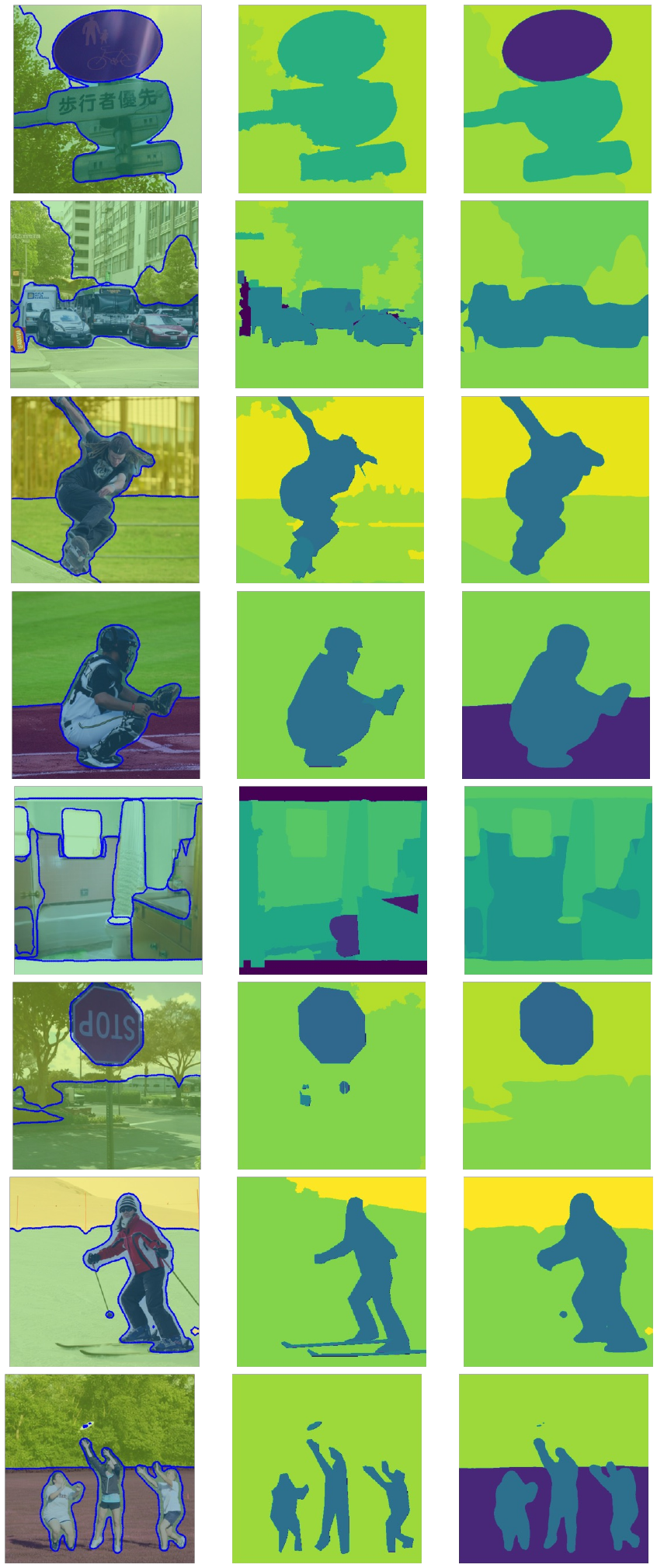}
    \caption{Examples of Segmentation on COCO-Stuff-27. Overlay (left),  ground truth (middle), and segmentation (right)}
    \label{fig:coco_seg}
\end{figure}
\begin{figure}[ht]
    \centering
    \includegraphics[width=0.5\textwidth]{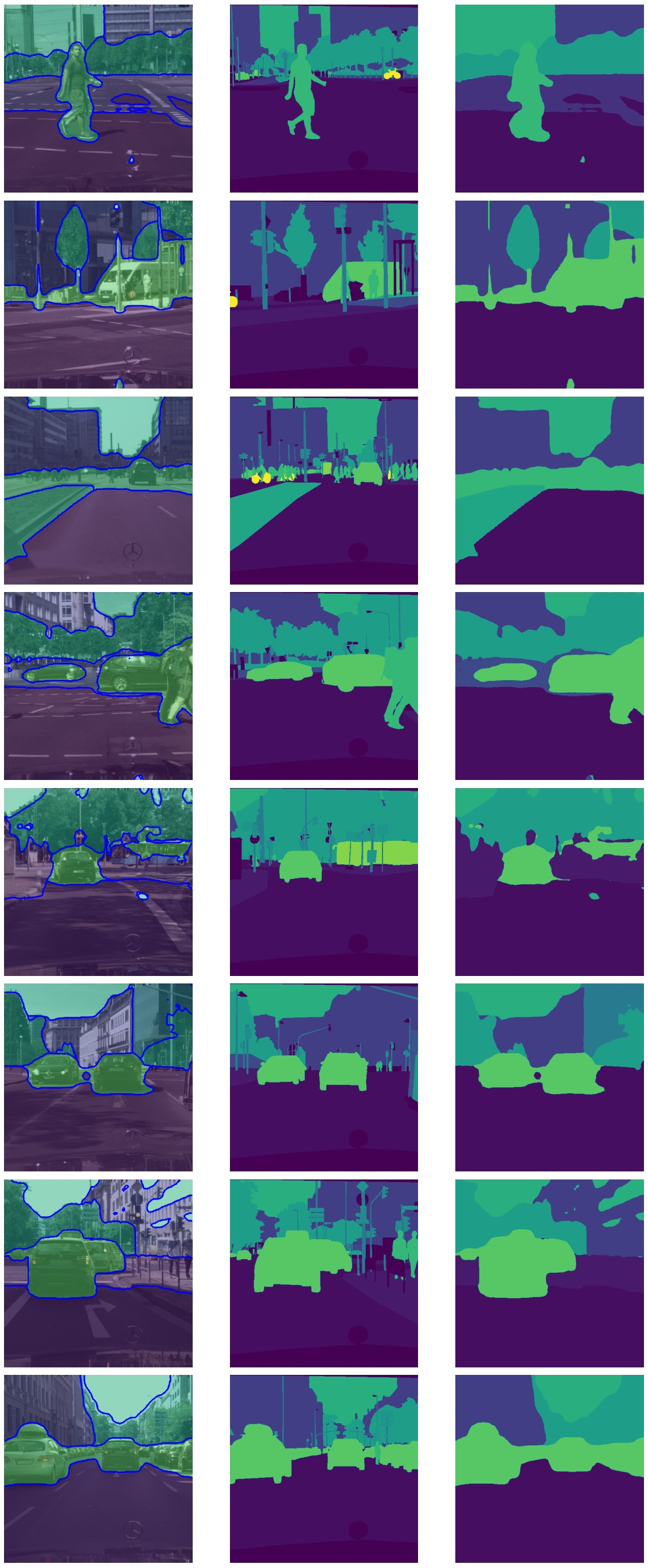}
    \caption{Examples of Segmentation on Cityscapes. Overlay (left), ground truth (middle), and segmentation (right)}
    \label{fig:cityscapes_seg}
\end{figure}
\begin{figure}[ht]
    \centering
    \includegraphics[width=0.5\textwidth]{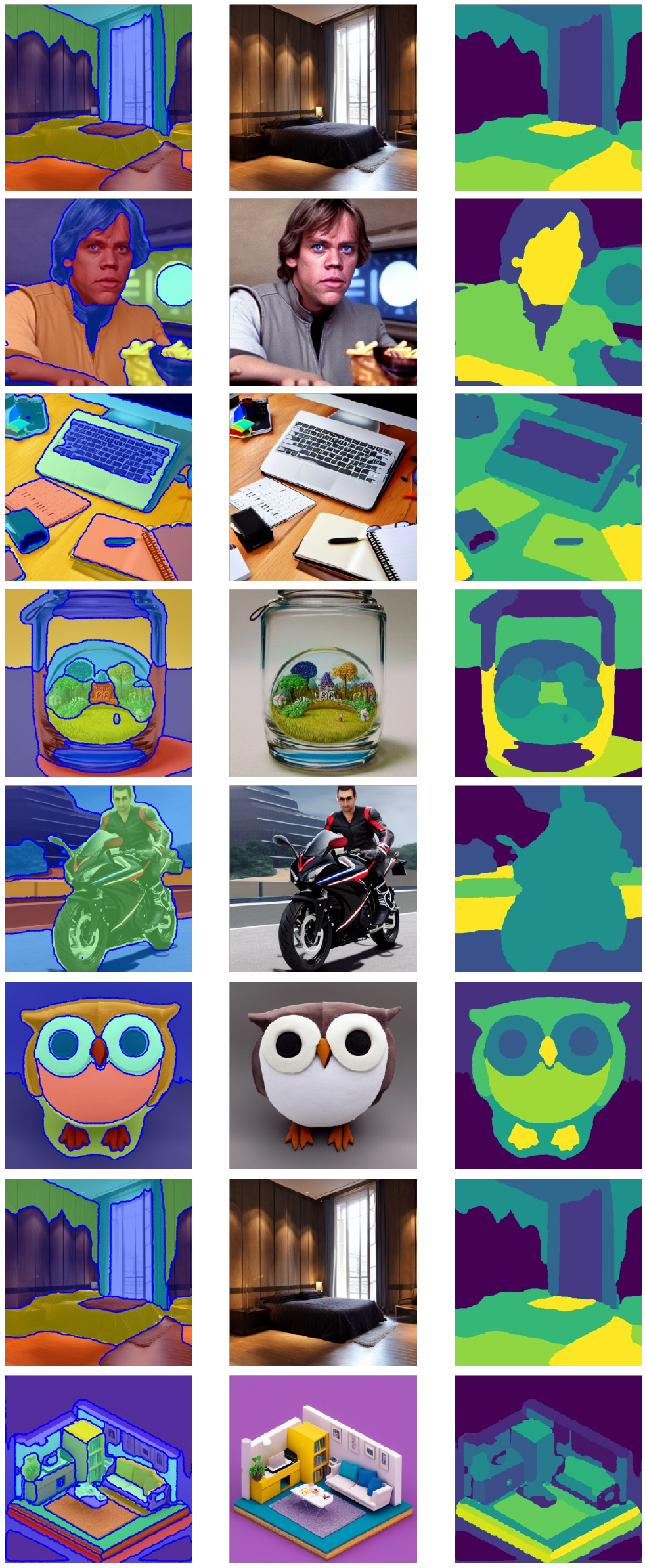}
    \caption{Examples of Segmentation on Synthetic Images (generated by a stable diffusion model). Overlay (left), Input (middle), and segmentation (right)}
    \label{fig:synthetic_seg}
\end{figure}

\clearpage
\newpage

\end{document}